\documentclass[lettersize,journal]{IEEEtran}
\usepackage{amsmath,amsfonts}
\usepackage{algorithm}
\usepackage{array}
\ifCLASSOPTIONcompsoc
    \usepackage[caption=false, font=normalsize, labelfont=sf, textfont=sf]{subfig}
\else

\usepackage[caption=false, font=footnotesize]{subfig}
\fi
\usepackage{textcomp}
\usepackage{stfloats}
\usepackage{url}
\usepackage{verbatim}
\usepackage{graphicx}
\usepackage{cite}
\usepackage{acronym}
\usepackage{algorithm}
\usepackage{algpseudocode}
\usepackage{bm}\usepackage{multirow}
\usepackage{color}
\usepackage{float}
\usepackage{lineno}
\usepackage{diagbox}
\acrodef{AWA}{Auxiliary Weights Adaptation}

\def \ip {image inpainting }
\def \Ip {Image inpainting }
\def \ea {et al. }

\def \L {\mathcal{L}}
\acrodef{TPL}{Tunable Perceptual Loss}
\acrodef{TSL}{Tunable Style Loss}

\newcommand{\Eqref}[1]{{Eq.~\ref{#1}}}
\newcommand{\note}[2]{\mathop{\underbrace{#1}}_{#2}}
\newcommand{\Figref}[1]{{Fig.~\ref{#1}}}
\newcommand{\Tabref}[1]{{TABLE~\ref{#1}}}

\def \The {\theta}
\def \Des {\mathbf{x}_{i}}
\def \Gt {\mathbf{y}_{i}}
\def \Pred {\tilde{\mathbf{y}}_{i}}

\def \lsw {\phi_s}

\newcommand{\de}[1] {{\nabla_{#1}}}

\def \is {\mathcal{L}_{c}}
\def \ic {\mathcal{L}_{t}}

\def \pp {\phi_{p}}

\renewcommand{\Omega}{\omega(\phi)}
\def\allfiles{}
\def\alltabs{}
\begin{document}
\title{Auxiliary Loss Reweighting for Image Inpainting}

\author{Siqi~Hui, Wenli~Huang, Sanping~Zhou, Ye~Deng, Jinjun~Wang
\thanks{Manuscript received July 21 2022. This work is jointly supported by the National Key Research and Development Program of China under Grant No. 2017YFA0700800, the General Program of China Postdoctoral Science Foundation under Grant No. 2020M683490, and the Youth program of Shaanxi Natural Science Foundation under Grant No. 2021JQ-054.}
\thanks{Siqi~Hui, Wenli~Huang, Sanping~Zhou, Ye~Deng, Jinjun~Wang are with Institute of Artificial Intelligence and Robotics, Xi'an Jiaotong University, Xi'an, Shanxi, 710049, China (huisiqi@stu.xjtu.edu.cn, 
huangwenwenlili@126.com,
spzhou@xjtu.edu.cn, 
dengye@stu.xjtu.edu.cn, 
jinjun@mail.xjtu.edu.cn).}
\thanks{Correspondingg author: Jinjun~Wang}}



\maketitle
\ifx\allfiles\undefined
\documentclass[lettersize,journal]{IEEEtran}
\usepackage{amsmath,amsfonts}
\usepackage{algorithm}
\usepackage{array}
\usepackage[caption=false,font=normalsize,labelfont=sf,textfont=sf]{subfig}
\usepackage{textcomp}
\usepackage{stfloats}
\usepackage{url}
\usepackage{verbatim}
\usepackage{graphicx}
\usepackage{cite}
\hyphenation{op-tical net-works semi-conduc-tor IEEE-Xplore}

\begin{document}
\linenumbers
\fi

\begin{abstract}
Image Inpainting is a task that aims to fill in missing regions of corrupted images with plausible contents. Recent inpainting methods have introduced perceptual and style losses as auxiliary losses to guide the learning of inpainting generators. Perceptual and style losses help improve the perceptual quality of inpainted results by supervising deep features of generated regions. However, two challenges have emerged with the usage of auxiliary losses: (i) the time-consuming grid search is required to decide weights for perceptual and style losses to properly perform, and (ii) loss terms with different auxiliary abilities are equally weighted by perceptual and style losses. To meet these two challenges, we propose a novel framework that independently weights auxiliary loss terms and adaptively adjusts their weights within a single training process, without a time-consuming grid search. Specifically, to release the auxiliary potential of perceptual and style losses, we propose two auxiliary losses, \ac{TPL} and \ac{TSL} by using different tunable weights to consider the contributions of different loss terms. TPL and TSL are supersets of perceptual and style losses and release the auxiliary potential of standard perceptual and style losses. We further propose the \ac{AWA} algorithm, which efficiently reweights TPL and TSL in a single training process. AWA is based on the principle that the best auxiliary weights would lead to the most improvement in inpainting performance. We conduct experiments on publically available datasets and find that our framework helps current SOTA methods achieve better results. 
\end{abstract}


\ifx\allfiles\undefined
\end{document}
\fi
\ifx\allfiles\undefined
\documentclass[lettersize,journal]{IEEEtran}
\usepackage{amsmath,amsfonts}
\usepackage{algorithm}
\usepackage{array}
\usepackage[caption=false,font=normalsize,labelfont=sf,textfont=sf]{subfig}
\usepackage{textcomp}
\usepackage{stfloats}
\usepackage{url}
\usepackage{verbatim}
\usepackage{graphicx}
\usepackage{cite}
\hyphenation{op-tical net-works semi-conduc-tor IEEE-Xplore}

\begin{document}
\linenumbers
\fi

\begin{figure}[htbp]
    \centering
    \subfloat[\label{1a}]{
       \includegraphics[width=1.\linewidth]{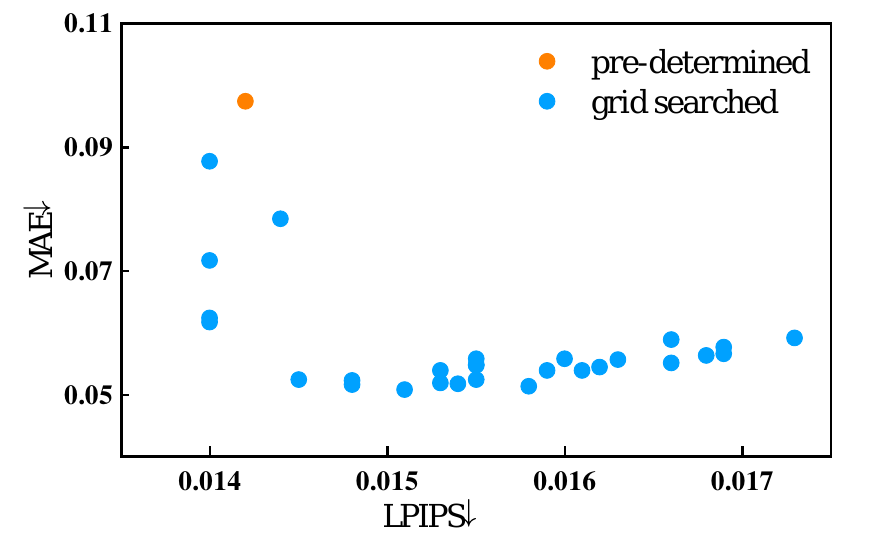}}\\
     \centering
     \subfloat[\label{1b}]{%
       \includegraphics[width=1.\linewidth]{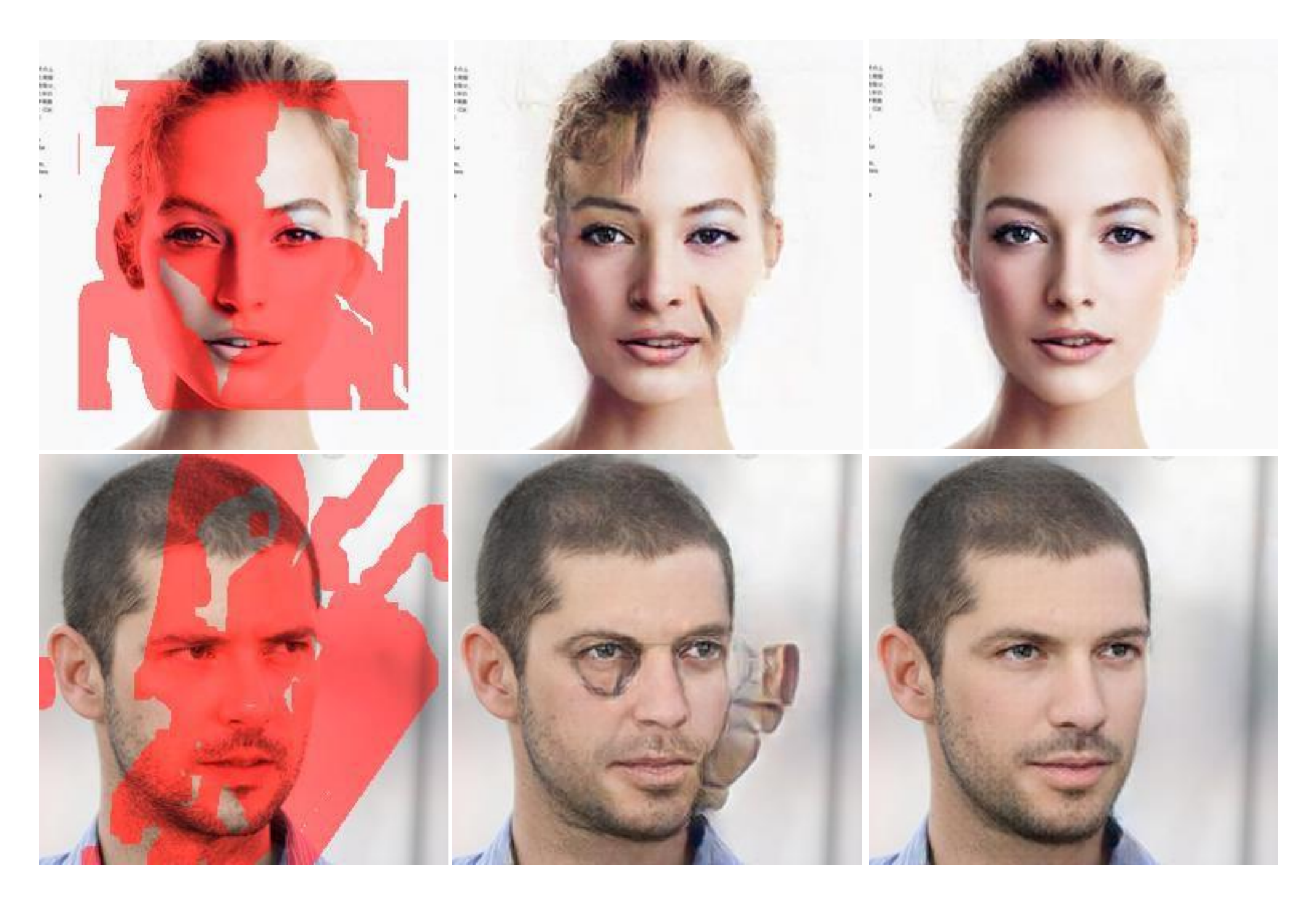}}
    \caption{(a) Inpainting performances of RFR~\cite{li2020recurrent} model trained with differently weighted perceptual and style losses on CelebA-HQ~\cite{celeba2018} dataset, and inpainting performance is sensitive to auxiliary weights. Blue dots represent models trained with grid searched auxiliary weights, and the orange dot is the model trained following default settings.  $^\downarrow$ means lower is better. (b) Visual results of default model and grid searched model. From left to right: masked images, default results and grid searched results. Better results can be found through grid search.}
    \label{fig: grid search}
\end{figure}

\section{Introduction}
Image inpainting is an image restoration task that aims to fill in corrupted image regions with visually-plausible contents coherent with untainted regions. \Ip is widely adopted in real-world applications, such as face editing~\cite{jo2019sc} , object removal~\cite{uittenbogaard2019privacy}, photo restoration etc. The challenge of the inpainting task is to generate patches with limited information from known regions.

Early \ip methods \cite{barnes2009patchmatch,drori2003fragment,Ballester2001Filling} restore missing regions by searching best patches based on hand-crafted features or propagation, which perform well when dealing with images with modest corruptions. Benefit from the development of deep neural networks, recent learning-based \ip methods~\cite{pathak2016context,gated2019,contextualattention2018,li2020recurrent,guo2021ctsdg} employed deep Convolutional Neural Networks (CNNs) to build generators that model distant relationships and generate high-quality patches for images with large holes, and the generators are guided by per-pixel reconstruction loss and adversarial loss~\cite{2017GANs}. Many current \ip methods \cite{wang2021dynamic,zhu2021image,guo2021ctsdg} also introduced perceptual loss~\cite{gatys2015neural} and style loss~\cite{imagenet2015} as auxiliary losses to further improve the perceptual quality of inpainting results. Standard perceptual or style loss is defined as the summation of several loss terms. Loss terms are based on different deep features of generated and ground truth images, where the deep features are extracted by pre-trained VGG network~\cite{simonyan2014very}. Different from reconstruction or adversarial loss which directly supervise the generated images, the auxiliary losses supervise deep features of generated images which indirectly improves the inpainting performance.

Though perceptual and style losses improve inpainting performance, they face several practical challenges. Firstly, extensive hyperparameter tuning and time-consuming cross-validation are required to determine proper weights for auxiliary weights losses to be properly performed, and the training cost grows exponentially with the number of auxiliary weights (see figure~\ref{fig: grid search}). Secondly, their functional form restricts auxiliary potentials of deep features: They simply sum several loss terms up and output one coherent loss which equally treats different loss terms, while more useful terms should be emphasized. These challenges motivate us to study the auxiliary loss reweighting algorithm which and release the full potential of feature-based auxiliary losses.

In this paper, we propose a novel framework to release the auxiliary potential of perceptual loss and style loss and eliminate hyperparameter tuning. We first propose \ac{TPL} and \ac{TSL} to enable an independent weighting paradigm. \ac{TPL} and \ac{TSL} use a group of trainable weights to consider contributions of different loss terms of perceptual and style losses, viewing each loss term as a separate auxiliary loss. They model a family of auxiliary losses for \ip which provides a large search space for the grid search. Then, to efficiently determine the proper weights for \ac{TPL} and \ac{TSL}, we propose \ac{AWA} algorithm which dynamically adjusts weights of \ac{TPL} and \ac{TSL} and achieves competitive inpainting results in a single training process. \ac{AWA} achieves this by providing an explicit optimization goal for auxiliary weights: maximizing the inpainting performance after several iterations of training steps. The optimization goal is based on the principle that the best weighted auxiliary losses would lead to the most pleasing inpainting performance. 
 
Experiments on four publicly available datasets demonstrate that best \ip performance can be achieved by our framework compared with other \ip methods. We further show that \ac{AWA} algorithm is compatible with current inpainting methods and is better than other loss reweighting competitors.
Our paper has the following three contributions:
\begin{itemize}
    \item We propose \ac{AWA} algorithm which automatically adjusts auxiliary weights. As far as we know, it is the first time that hyperparameter learning mechanism is introduced to image inpainting. 
    \item We also propose TPL and TSL which model a large family of auxiliary losses and increase the effectiveness of standard perceptual loss and style loss. 
    \item Experiments on four publicly available datasets demonstrate the superiority and universality of our framework.
\end{itemize}

The rest of the paper is organized as follows. Section II summarizes the related works on image inpainting and loss weights adaptation methods. Section III illustrates our framework, including the form of \ac{TPL} and \ac{TSL} losses and the \ac{AWA} algorithm. In section IV, we discuss the experimental settings. Section V shows the experimental results. In the last section, we conclude the paper and propose future work.
\ifx\allfiles\undefined
\bibliographystyle{IEEEtran}
\bibliography{reference/reference.bib,reference/diffusion_based.bib,reference/patch_based.bib,reference/diffusion_patch_based.bib,reference/loss_reweighting_alg.bib}
\end{document}
\fi
\ifx\allfiles\undefined
\documentclass[lettersize,journal]{IEEEtran}
\usepackage{amsmath,amsfonts}
\usepackage{algorithm}
\usepackage{array}
\usepackage[caption=false,font=normalsize,labelfont=sf,textfont=sf]{subfig}
\usepackage{textcomp}
\usepackage{stfloats}
\usepackage{url}
\usepackage{verbatim}
\usepackage{graphicx}
\usepackage{cite}
\hyphenation{op-tical net-works semi-conduc-tor IEEE-Xplore}

\begin{document}
\linenumbers
\fi

\section{Related Work}
\subsection{Image Inpainting}
The image inpainting approaches are mainly branched into the following groups: patch-based image inpainting \cite{barnes2009patchmatch,kwatra2005texture,drori2003fragment}, diffusion-based image inpainting \cite{barbu2016variational,bertozzi2011unconditionally,chan2001nontexture,shen2003euler,sridevi2019image,Ballester2001Filling}, and learning-based image inpainting~\cite{pathak2016context,contextualattention2018,gated2019,liu2018partial,nazeri2019edgeconnect,li2019progressive,yeh2017semantic,lahiri2020prior,li2020recurrent,liu2021pd,yi2020contextual,wang2021dynamic,song2018spg,liu2019coherent,suin2021distillation,xiong2019foreground,yu2021wavefill,cao2021sketch,zhu2021image}.

Patch-based methods synthesize the unknown regions by the most identical matched patches in a known area. An influential patch-based inpainting approach was developed by Barnes et al. \cite{barnes2009patchmatch}. Diffusion-based methods fill in tiny missing regions by propagation to diffuse information from the outside to the inside of the hole region. Some other methods~\cite{ding2018image,ding2018perceptually} combine diffusion-based and patch-based techniques to generate better results. However, these methods have limited performance when dealing with large holes, as they only consider low-level features of the background.

Learning-based methods could generate deterministic high-quality results with low distortion. They use CNN to extract high-level features and model long-distance relationships between the missing and known region. Some methods~\cite{liu2018partial,gated2019,wang2021dynamic,su2019pixel,suin2021distillation} design specific convolution layers for \ip. Some try to use structure information (contour \cite{xiong2019foreground}, canny edge \cite{nazeri2019edgeconnect}, gradient map \cite{yang2020learning}, segmentation map \cite{song2018spg,liu2019coherent,lahiri2020prior}) to assist inpainting process. Some other methods~\cite{yeh2017semantic,iizuka2017globally,pathak2016context,yu2020region,yi2020contextual,liu2021pd} introduce adversarial loss to increase and diversity of generated results.

Auxiliary losses (e.g. perceptual loss~\cite{gatys2015neural} and style losses~\cite{liu2018partial}) are further introduced to increase perceptual quality. Based on deep features, the perceptual loss is used in style transfer learning \cite{gatys2015neural} and was firstly introduced by~\cite{liu2018partial} to handle image inpainting. Style loss is used to ameliorate the periodic unreal patterns brought by perceptual loss. Perceptual and style losses are different from MAE or adversarial losses, as they measure the difference between deep features of generated and ground truth images. Though the perceptual and style losses improve inpainting performance, their functional form restricts auxiliary abilities of deep features, as loss terms are equally treated by simply summing several loss terms up. In this paper, we attempt to independently weight loss terms and dynamically adjust their weights. 

\subsection{Loss Reweighting Algorithms}
The correlated work to ours are loss reweighting algorithms that adjust loss weights to achieve balance among multiple loss functions. 

In the context of Multi-task learning (MTL), models are trained to give high performance on several different tasks~\cite{caruana1997multitask, thrun2012learning, ruder2017overview}. If several related tasks could assist each other by sharing certain parameters~\cite{ando2005framework,sener2018multi}, better generalization can be achieved~\cite{girshick2015fast,sharma2017online}. To achieve better generalization, MTL methods employ loss reweighting algorithms to balance multiple losses. Sener~\ea~\cite{sener2018multi} adjusts the combination among tasks by minimizing the upper bound of the multiple-gradient descent algorithm (MGDA-UB). Chen~\ea~\cite{chen2018gradnorm} adapts task weights by manually restricting gradient norms of different tasks on a common scale to learn tasks at the same pace. At the same time, Kendall~\ea~\cite{kendall2018multi} proposed a principled approach that adjusts weights according to uncertainty of tasks. To achieve a better trade-off between accuracy and computational cost of anytime neural networks (ANN)~\cite{huang2017multi}, Hu~\ea~\cite{hu2019learning} makes the weights are inversely proportional to the average of each loss. 

Closely related to MTL, a good balance between main and auxiliary losses need to be achieved in auxiliary learning (AL) where main and auxiliary losses are jointly optimized, but only the main task's performance is important. To better use auxiliary losses, in \cite{du2018gradsim} an auxiliary loss is filtered out when it has an opposite gradient direction to the main loss. Some other algorithms have been designed to adjust auxiliary weights by estimating noisy gradient similarity (e.g. inner product \cite{lin2019adaptive} and the l2 distance \cite{shi2020auxiliary}) between main and auxiliary losses. Different from these works, we provide the stable and explicit optimization goal for auxiliary weights. 

It is noteworthy that, the proposed \ac{AWA} is different from current adaptive auxiliary loss techniques \cite{hu2019learning,du2018gradsim,chen2018gradnorm} which implicitly adjust auxiliary weights by considering loss magnitudes or noisy gradient similarities among main loss and auxiliary losses. The \ac{AWA} provides an explicit optimization goal for auxiliary weights and increases the inpainting performance.
\ifx\allfiles\undefined
\bibliographystyle{IEEEtran}
\bibliography{reference/reference.bib,reference/diffusion_based.bib,reference/patch_based.bib,reference/diffusion_patch_based.bib,reference/loss_reweighting_alg.bib}
\end{document}
\fi
\ifx\allfiles\undefined
\documentclass[lettersize,journal]{IEEEtran}
\usepackage{amsmath,amsfonts}
\usepackage{float}
\usepackage{lineno}
\usepackage{algorithm}
\usepackage{array}
\usepackage[caption=false,font=normalsize,labelfont=sf,textfont=sf]{subfig}
\usepackage{textcomp}
\usepackage{stfloats}
\usepackage{url}
\usepackage{verbatim}
\usepackage{graphicx}
\usepackage{cite}
\hyphenation{op-tical net-works semi-conduc-tor IEEE-Xplore}

\begin{document}
\fi
\section{Approach}
In this section, we first define the notation and the inpainting problem and describe details of \ac{TPL} and \ac{TSL}, then we propose our \ac{AWA} algorithm to dynamically reweight them by setting an explicit optimization goal for weights of \ac{TPL} and \ac{TSL}.

\subsection{Notations and Definitions}
Let $\{(\Des,\Gt)\}_i$ be the training dataset for image inpainting, where $\Des$ and $\Gt$ are the corrupted image and ground truth image, respectively. Current inpainting methods try to learn parameterized generators $g(\cdot;\The)$ to approximate the function which maps destroyed images to corresponding ground truth. Specifically, after receiving a corrupted image $\Des$, the generator predicts the corresponding ground truth $\Pred=g(\Des;\The)$.  

The inpainting loss can be defined as the summation of a coherent main loss and an auxiliary loss:
\begin{equation}
\begin{aligned}
\label{eq:trainloss}
&\ic(\theta,\phi)=\lambda_m\cdot L_{m}(\The) + \lambda_a\cdot L_a(\The),\\
&L_m(\The)=(\L_{l1}(\The),\L_{adv}(\The),\L_{tv}(\The))^T,\\
&L_a(\The)=(\L_{perc}(\The),\L_{styl}(\The))^T,
\end{aligned}
\end{equation}
where the main loss is the dot product of loss vector $L_m$ and weight vector $\lambda_m$. $L_m$ is comprised of the $l_1$, adversarial and total variance losses. The coherent auxiliary loss is comprised of perceptual and style losses and weight vector $\lambda_a$. The forms of perceptual loss $\L_{perc}$ and style loss $\L_{styl}$ are as follows: 
\begin{equation}
 \label{ploss}
\begin{aligned}
     &\L_{perc}(\The)= \sum_{n} \L_{perc}^n(\The),\qquad\qquad\\
    &\L_{perc}^n(\The)=\sum_i ||\bm{\psi}_{n}(\Pred)-\bm{\psi}_{n}(\Gt)||_{1},
\end{aligned}
\end{equation}

\begin{equation}
\label{sloss}
\begin{aligned}
&\L_{styl}(\The)= \sum_{n} \L_{styl}^n(\The),\\
&\L_{styl}^n(\The)=\sum_i ||G(\bm{\psi}_{n}(\Pred))-G(\bm{\psi}_{n}(\Gt))||_{1},
\end{aligned}
\end{equation}
where $\bm{\psi}$ denotes pretrained VGG parameters, and $\bm{\psi}_n(.)$ outputs the feature map of the $n$-th chosen layer of the VGG network. As shown in \Eqref{ploss}, perceptual loss measures the difference between deep features of generated and ground truth images. Style loss has a similar form, except that it measures the Gram matrices. In \Eqref{sloss} $G(.)$ outputs the Gram matrix of the input deep feature.
 
\subsection{Tunable Perceptual and Style Losses}
Perceptual and style losses have been introduced by current \ip methods \cite{liu2018partial,nazeri2019edgeconnect,li2019progressive,wang2021dynamic,li2020recurrent,guo2021ctsdg} to improve the perceptual quality. As mentioned above, they add several loss terms up and use the same weight considering the contribution of different loss terms. However, it restricts the auxiliary potential of deep features, as more useful terms should be emphasized. 

To release the potential of deep feature based auxiliary losses, we propose tunable perceptual and style losses (\ac{TPL} \& \ac{TSL}) which use a group of continuous-valued weights to independently consider contributions of loss terms of perceptual and style losses. In effect, the \ac{TPL} takes the following form:
\begin{equation}
\begin{aligned}
    \label{eq:tunable perceptual loss} 
    &\mathcal{L}_{tperc}(\The) = \omega_p(\phi_p) \cdot L_{p}(\The),\\
    &L_{p}(\The)=(\L_{perc}^1(\The),...,\L_{perc}^N(\The))^T,\\
    &\omega_p(\phi_p)=\lambda_p \odot \delta(\phi_p).
\end{aligned}
\end{equation}
As shown in \Eqref{eq:tunable perceptual loss}, \ac{TPL} is the weighted sum of N perceptual loss terms $L_p$. The weight vector $\omega_p(\phi_p)$ is the function of continuous-valued parameters $\phi_p$. $\psi_p$ is activated by sigmoid $\delta(.)$ and element-wise rescaled by $\lambda_p$ such that the value range of weights of \ac{TPL} is from 0 to $\lambda_p$, where $\odot$ represents the Hadamard product. Note that TPL is a superset of perceptual loss, and in experiments, all elements of $\psi_p$ are equally initialized such that TPL equals perceptual loss to make fair comparisons. Following \ac{TPL}, we propose \ac{TSL} to model a family of style loss by adding $\lsw$ to standard style loss, and \ac{TSL} is defined as follows:
\begin{equation}
\begin{aligned}
    \label{eq:tunable style loss} 
    &\mathcal{L}_{tstyl}(\The) = \omega_s(\phi_p) \cdot L_{s}(\The),\\
    &L_{s}(\The)=(\L_{styl}^1(\The),...,\L_{styl}^N(\The))^T,\\
    &\omega_s(\phi_p)=\lambda_s \odot \delta(\phi_s).
\end{aligned}
\end{equation}
where the sigmoid is also applied on $\phi_s$, and $\lambda_s$ denotes the maximum value of weights of \ac{TSL}. To simplify notations, we use $\omega_p$ and $\omega_s$ to represent $\omega_p(\phi_p)$ and $\omega_s(\phi_s)$ respectively. Then the perceptual and style losses can be replaced with \ac{TPL} and \ac{TSL}, and the new inpainting loss is defined as follows:
\begin{equation}
\label{eq:total_loss2}
\begin{aligned}
    \ic(\theta)=&\lambda_m\cdot L_{m}(\The) + \omega_p\cdot L_{p}(\The)+\omega_s\cdot L_{s}(\The).
\end{aligned}
\end{equation}

\subsection{Auxiliary Weights Adaptation}
 As mentioned above, the \ac{TPL} and \ac{TSL} model a larger family of auxiliary loss for image inpainting. However, it is impractical to use grid search to determine proper weights for \ac{TPL} and \ac{TSL}, as it takes days to train a single inpainting model with one combination of auxiliary weights. To properly reweight auxiliary losses, we propose \ac{AWA} algorithm which alters between optimizing model parameters ($\The$) and adaptively adjusting auxiliary parameters ($\phi_p$ and $\phi_s$) in a single training process. 
 
Auxiliary parameters are learned by optimizing the optimizing goal: maximizing the inpainting performance measured by some metric (e.g. LPIPS) after the model trained by the \ac{TPL} and \ac{TSL} for $K$ steps. Formally, objective of auxiliary parameters $\phi_p$ and $\phi_s$ is as follows:
\begin{equation}
\begin{aligned}
\label{phiobj}
    \phi_p^*,\phi_s^*=&\mathop{argmin}_{\phi_p,\phi_s} \is(\The^K), \\
    s.t. \ \The^K=&\The-\alpha\sum_{j=1}^K(\lambda_m^T J_m(\The^j)+\omega_p^T J_p(\The^j)\\ &+\omega_s^T J_s(\The^j)),
\end{aligned}
\end{equation}
where the metric is defined as $\is(\The^K)=\sum_i l_c(\Des,\Gt,\The^K)$. $\The^K$ denotes the model parameter updated $K$ steps from $\The$, and $\alpha$ is the model learning rate. $J_m(\The^j)$, $J_p(\The^j)$ and $J_s(\The^j)$ are Jacobian matrices of vector loss $L_m$, $L_p$ and $L_s$ with respect to generator parameter at $j$-th step (In Adam~\cite{kingma2014adam} these matrices are also rescaled element-wise).

Though we can directly optimize auxiliary parameters on $\is$ through a gradient-based process, it requires to save $K$ steps of computational graph and differentiating through the optimization process over $\The$. In practice, we solve this by proposing the surrogate loss function $\L_{sg}(\phi_p,\phi_s)$, which takes the following form:
\begin{equation}
    \label{eq:lossaux}
\begin{aligned}
\L_{sg}(\phi_p,\phi_s)=&-\nabla_\The^T\L_c(\The^K) \sum_j (J_p^T(\The^j)\omega_p
\\ &+J_s^T(\The^j)\omega_s).
\end{aligned}
\end{equation}
The surrogate loss is comprised of Jacobian matrices of auxiliary losses and the gradient of $\is$ at point $\The^K$ without differentiating through K step updates of model parameters. Please refer to the Appendix for the derivation of \Eqref{eq:lossaux}. 

After the update of auxiliary parameters, model parameters $\The$ could be learned by newly weighted auxiliary losses using the loss function in \Eqref{eq:total_loss2}. Our algorithm is summarized in Algorithm.\ref{alg:Framwork}. We show a gradient descent version of \ac{AWA} for simplicity, and parameters can also be optimized by other optimizers such as Adam. In practice, we use LPIPS as the guiding metrics, as the ablative studies show that LPIPS is more fit for \ac{AWA}, please refer to section~\ref{ablative} for more details.

\begin{algorithm}[t]
\caption{Adaptive Auxiliary Loss.} 
\label{alg:Framwork}
\begin{algorithmic} 
\Require
Pre-determined hyperparameter $K$, $\lambda_p$ and $\lambda_s$. Randomly initialized auxiliary parameters $\phi_p$ and $\phi_s$ and the learning rate $\beta$. Randomly initialized model parameters $\The$ and the corresponding learning rate $\alpha$.
\While {$\theta \ not \ converged$}
    \State Compute $\The^K$: $\The^K=\The-\alpha\sum_{j=1}^K(\lambda_m^T J_m(\The^j)+\omega_p^T J_p(\The^j)$
    \State ~~~~~~~~~~~~~~~~~~~~~~~$+\omega_s^T J_s(\The^j))$
    \State Save Jacobian matrices: \State $\{J_p(\The^1),...,J_p(\The^K),J_s(\The^1),...,J_s(\The^K)\}$
    \State Train auxiliary parameters with loss function \Eqref{eq:lossaux}:
    \State $\phi_p\leftarrow\phi_p-\beta\nabla_{\phi_p}\L_{sg}(\phi_p,\phi_s)$
    \State $\phi_s\leftarrow\phi_s-\beta\nabla_{\phi_s}\L_{sg}(\phi_p,\phi_s)$
    \State Train $\The$ with reweighted \ac{TPL} and \ac{TSL} with \Eqref{eq:total_loss2}:
    \State $\The\leftarrow\The-\alpha( J_m(\The^j)\lambda_m+J_p(\The^j)\omega_p+J_s(\The^j)\omega_s))$
\EndWhile
\end{algorithmic} 
\end{algorithm}


\ifx\allfiles\undefined
\bibliographystyle{IEEEtran}
\bibliography{reference/reference.bib}
\end{document}
\fi
\ifx\allfiles\undefined
\documentclass[lettersize,journal]{IEEEtran}
\usepackage{amsmath,amsfonts}
\usepackage{algorithm}
\usepackage{array}
\usepackage[caption=false,font=normalsize,labelfont=sf,textfont=sf]{subfig}
\usepackage{textcomp}
\usepackage{stfloats}
\usepackage{url}
\usepackage{verbatim}
\usepackage{graphicx}
\usepackage{cite}
\hyphenation{op-tical net-works semi-conduc-tor IEEE-Xplore}

\def\alltabs{}
\begin{document}
\fi

\section{Experimental Settings}
\subsection{Dataset}
We conduct experiments on four commonly used datasets in experiments, namely CelebA \cite{celeba2018}, CelebA-HQ, Paris-StreetView \cite{paris2012} and Places2 \cite{zhou2017places}. Besides, an external mask dataset from \cite{liu2018partial} is also used to produce corrupted images.
\begin{itemize}
    \item \emph{CelebA}: Celeba contains more than 200,000 face images, include 162770 training images, 19867 validation images and 19962 testing images, we randomly select 10,000 images from test dataset for testing.
    \item \emph{CelebA-HQ}: CelebA-HQ is a dataset that has 30,000 high-resolution face images selected from the CelebA dataset, include 28000 training images and 2000 testing images. All images are resized to $256 \times 256$ in our experiments.
    \item \emph{Paris StreetView}: Paris StreetView contains 15,000 outdoor building images, with 14900 images for training and 100 images for testing.
    \item \emph{Places2}: Places2 is a collection 365 categories scene images which contains 1803640 training images, and we randomly select 10000 images in the test dataset for testing.
    \item \emph{Masks}: We use irregular mask dataset provided by~\cite{liu2018partial}, in which masks are split into groups according to the relative masked ratio: (0.01,0.1], (0.1,0.2], (0.3,0.4], (0.4,0.5], (0.5,0.6]. Each group has 2000 images.
\end{itemize}

The size of the training image is 256 × 256. For CelebA, the raw images have a size of 178 × 218, and we crop the 178 x 178 region at the center and resize it to 256 x 256 as the input image. For Paris StreetView and Places2, resizing operation is only needed, as images are squre. 
\\

\subsection{Inpainting Methods}
Our framework is compatible with current CNN based inpainting methods, and in this paper, we apply our framework to train the following inpainting models. 
\begin{itemize}
    \item \emph{EG}\cite{nazeri2019edgeconnect} EG uses a two-stage model where the first generator completes edge maps and image generator uses these maps as structure priors to generate images.
    \item \emph{PRVS}\cite{li2019progressive} PRVS devises a network which progressively reconstructs strucure maps as well as visual features.
    \item \emph{CTSDG}\cite{guo2021ctsdg} CTSDG designs a two-stream network which utilizes texture and structure reconstruction to guide the training of encoder and decoder respectively.
\end{itemize}

\subsection{Implementation Details}
Our framework are applied to train various SOTA methods~\cite{guo2021ctsdg,nazeri2019edgeconnect,li2019progressive}. 
All experiments are implemented by Python on Ubuntu 20.04 system with 2 GeForce RTX 3090 GPUs. With \ac{AWA}, models are trained following the settings of the SOTA methods. The auxiliary parameters are updated by the AdamW optimizer with $\beta_1=0.5$ and $\beta_2=0.999$, and the learning rate $\beta$ is $1e^{-3}$. The hyperparameter $K$ is set to 1.  For \ac{TPL} and \ac{TSL}, $\lambda_p$ and $\lambda_s$ are set to 2 and 750 respectively. Analogous to~\cite{guo2021ctsdg}, the deep features of \ac{TPL} and \ac{TSL} are extracted by deep features are from the first three max-pooling layers of the pre-trained VGG-16 network. Our source code is available at https://github.com/HuiSiqi/Auxiliary-Loss-Reweighting-for-Image-Inpainting.

\begin{figure*}[t]
    \centering
    \centerline{\includegraphics[width=1.\linewidth]{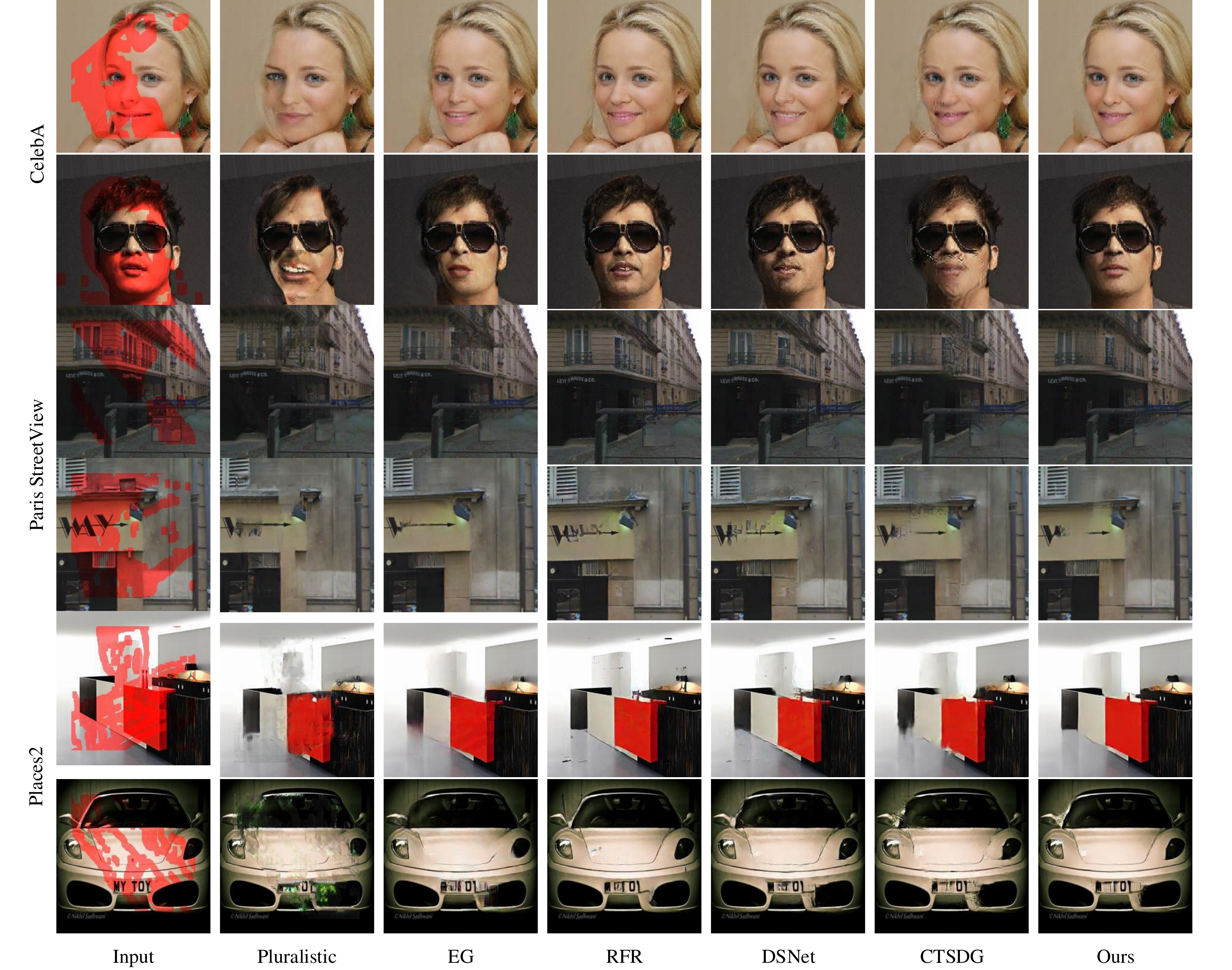}}
    \caption{Qualitative comparisons of our results and other \ip methods on CelebA, Paris StreetView and Places2. Here, we apply our framework to train CTSDG and generate our results.}
    \label{fig:comp-methods}
\end{figure*}

\ifx\alltabs\undefined
\documentclass[lettersize,journal]{IEEEtran}
\usepackage{amsmath,amsfonts}
\usepackage{algorithmic}
\usepackage{algorithm}
\usepackage{array}
\usepackage[caption=false,font=normalsize,labelfont=sf,textfont=sf]{subfig}
\usepackage{textcomp}
\usepackage{stfloats}
\usepackage{url}
\usepackage{verbatim}
\usepackage{graphicx}
\usepackage{cite}
\hyphenation{op-tical net-works semi-conduc-tor IEEE-Xplore}

\begin{document}
\fi

\begin{table*}[thbp]
\caption{Quantitative Comparisons of Our Results and Other Inpainting Methods. The Best Results Are \textbf{Boldfaced}. $^\uparrow$ Means Higher Is Better. $^\downarrow$ Means Lower Is Better.\label{tab:effectiveness}}
\centering
\begin{tabular}{p{0.07\textwidth}|p{0.07\textwidth}|p{0.07\textwidth}|p{0.07\textwidth}p{0.07\textwidth}p{0.07\textwidth}p{0.07\textwidth}p{0.07\textwidth}p{0.07\textwidth}p{0.08\textwidth}}
\hline
DataSet                                                                           & Metrics                              & Mask        & PIC    & EG     & PRVS    & RFR             & DSNet           & CTSDG  & Ours            \\ \hline
\multirow{20}{*}{CelebA}                                                          & \multirow{4}{*}{PSNR $^\uparrow$}    & (0.2,0.3{]} & 25.764 & 27.691 & 28.633  & 29.013          & 28.778          & 28.521 & \textbf{29.610} \\
                                                                                  &                                      & (0.3,0.4{]} & 23.528 & 26.368 & 26.373  & 26.687          & 26.439          & 26.200 & \textbf{27.202} \\
                                                                                  &                                      & (0.4,0.5{]} & 21.562 & 24.402 & 24.525  & 24.856          & 24.565          & 24.356 & \textbf{25.261} \\
                                                                                  &                                      & (0.5,0.6{]} & 18.457 & 21.398 & 21.750  & 22.057          & 21.776          & 21.610 & \textbf{22.294} \\ \cline{2-10} 
                                                                                  & \multirow{4}{*}{SSIM $^\uparrow$}    & (0.2,0.3{]} & 0.8984 & 0.9328 & 0.9417  & 0.9456          & 0.9433          & 0.9414 & \textbf{0.9518} \\
                                                                                  &                                      & (0.3,0.4{]} & 0.8568 & 0.9126 & 0.9113  & 0.9159          & 0.9124          & 0.9090 & \textbf{0.9240} \\
                                                                                  &                                      & (0.4,0.5{]} & 0.8078 & 0.8760 & 0.8767  & 0.8825          & 0.8772          & 0.8728 & \textbf{0.8921} \\
                                                                                  &                                      & (0.5,0.6{]} & 0.7198 & 0.8072 & 0.8148  & 0.8213          & 0.8137          & 0.8093 & \textbf{0.8329} \\ \cline{2-10} 
                                                                                  & \multirow{4}{*}{MAE $^\downarrow$}   & (0.2,0.3{]} & 0.0252 & 0.0136 & 0.0120  & 0.0111          & 0.0114          & 0.0115 & \textbf{0.0100} \\
                                                                                  &                                      & (0.3,0.4{]} & 0.0330 & 0.0175 & 0.0179  & 0.0168          & 0.0174          & 0.0177 & \textbf{0.0155} \\
                                                                                  &                                      & (0.4,0.5{]} & 0.0431 & 0.0249 & 0.0249  & 0.0234          & 0.0243          & 0.0248 & \textbf{0.0220} \\
                                                                                  &                                      & (0.5,0.6{]} & 0.0657 & 0.0404 & 0.0388  & 0.0366          & 0.0381          & 0.0391 & \textbf{0.0354} \\ \cline{2-10} 
                                                                                  & \multirow{4}{*}{LPIPS $^\downarrow$} & (0.2,0.3{]} & 0.0948 & 0.0637 & 0.0678  & \textbf{0.0465} & 0.0472          & 0.0659 & 0.0521          \\
                                                                                  &                                      & (0.3,0.4{]} & 0.1326 & 0.0822 & 0.1034  & \textbf{0.0710} & 0.0712          & 0.1005 & 0.0799          \\
                                                                                  &                                      & (0.4,0.5{]} & 0.1765 & 0.1156 & 0.1435  & \textbf{0.0987} & 0.0987          & 0.1389 & 0.1112          \\
                                                                                  &                                      & (0.5,0.6{]} & 0.2492 & 0.1766 & 0.2138  & 0.1474          & \textbf{0.1458} & 0.2044 & 0.1675          \\ \cline{2-10} 
                                                                                  & \multirow{4}{*}{FID $^\downarrow$}   & (0.2,0.3{]} & 5.2152 & 3.2487 & 4.4684  & 3.3639          & 2.9046          & 4.7415 & \textbf{2.7465} \\
                                                                                  &                                      & (0.3,0.4{]} & 7.5484 & 5.2663 & 8.7892  & 6.4810          & 5.4184          & 9.3803 & \textbf{4.6354} \\
                                                                                  &                                      & (0.4,0.5{]} & 10.832 & 8.8076 & 15.3408 & 11.038         & 9.0213          & 15.638 & \textbf{7.3770} \\
                                                                                  &                                      & (0.5,0.6{]} & 17.175 & 16.396 & 27.478  & 19.878          & 15.877          & 25.199 & \textbf{13.087} \\ \hline
\multirow{20}{*}{\begin{tabular}[c]{@{}c@{}}Paris\\      StreetView\end{tabular}} & \multirow{4}{*}{PSNR $^\uparrow$}    & (0.2,0.3{]} & 25.677 & 27.929 & 28.354  & 28.391          & 28.256          & 29.276 & \textbf{29.942} \\
                                                                                  &                                      & (0.3,0.4{]} & 23.698 & 25.817 & 26.241  & 26.310          & 26.158          & 26.888 & \textbf{27.664} \\
                                                                                  &                                      & (0.4,0.5{]} & 21.804 & 24.149 & 24.126  & 24.393          & 24.300          & 24.969 & \textbf{25.554} \\
                                                                                  &                                      & (0.5,0.6{]} & 18.942 & 21.747 & 21.419  & 21.867          & 21.763          & 22.263 & \textbf{22.502} \\ \cline{2-10} 
                                                                                  & \multirow{4}{*}{SSIM $^\uparrow$}    & (0.2,0.3{]} & 0.8698 & 0.9167 & 0.9233  & 0.9240          & 0.9210          & 0.9346 & \textbf{0.9418} \\
                                                                                  &                                      & (0.3,0.4{]} & 0.8180 & 0.8763 & 0.8835  & 0.8852          & 0.8802          & 0.8945 & \textbf{0.9059} \\
                                                                                  &                                      & (0.4,0.5{]} & 0.7519 & 0.8271 & 0.8315  & 0.8356          & 0.8302          & 0.8472 & \textbf{0.8609} \\
                                                                                  &                                      & (0.5,0.6{]} & 0.6475 & 0.7545 & 0.7446  & 0.7548          & 0.7488          & 0.7648 & \textbf{0.7807} \\ \cline{2-10} 
                                                                                  & \multirow{4}{*}{MAE $^\downarrow$}   & (0.2,0.3{]} & 0.0281 & 0.0155 & 0.0173  & 0.0171          & 0.0174          & 0.0125 & \textbf{0.0115} \\
                                                                                  &                                      & (0.3,0.4{]} & 0.0363 & 0.0222 & 0.0235  & 0.0232          & 0.0237          & 0.0189 & \textbf{0.0173} \\
                                                                                  &                                      & (0.4,0.5{]} & 0.0481 & 0.0307 & 0.0320  & 0.0312          & 0.0315          & 0.0266 & \textbf{0.0248} \\
                                                                                  &                                      & (0.5,0.6{]} & 0.0707 & 0.0444 & 0.0477  & 0.0454          & 0.0455          & 0.0409 & \textbf{0.0398} \\ \cline{2-10} 
                                                                                  & \multirow{4}{*}{LPIPS $^\downarrow$} & (0.2,0.3{]} & 0.1386 & 0.0778 & 0.1133  & 0.0696          & 0.0785          & 0.0697 & \textbf{0.0645} \\
                                                                                  &                                      & (0.3,0.4{]} & 0.1855 & 0.1129 & 0.1455  & \textbf{0.1006} & 0.1103          & 0.1109 & 0.1036          \\
                                                                                  &                                      & (0.4,0.5{]} & 0.2528 & 0.1608 & 0.1931  & \textbf{0.1413} & 0.1528          & 0.1636 & 0.1532          \\
                                                                                  &                                      & (0.5,0.6{]} & 0.3441 & 0.2298 & 0.2750  & \textbf{0.2065} & 0.2204          & 0.2471 & 0.2440          \\ \cline{2-10} 
                                                                                  & \multirow{4}{*}{FID $^\downarrow$}   & (0.2,0.3{]} & 71.772 & 41.273 & 35.892  & 31.319          & 35.758          & 39.695 & \textbf{30.985} \\
                                                                                  &                                      & (0.3,0.4{]} & 86.397 & 53.215 & 48.371  & 40.609          & 46.183          & 52.841 & \textbf{39.842} \\
                                                                                  &                                      & (0.4,0.5{]} & 101.88 & 70.144 & 68.103  & 55.892          & 61.823          & 75.214 & \textbf{53.016} \\
                                                                                  &                                      & (0.5,0.6{]} & 117.73 & 89.791 & 99.863  & 73.472          & 83.765          & 99.140 & \textbf{76.936} \\ \hline
\multirow{20}{*}{Places2}                                                         & \multirow{4}{*}{PSNR $^\uparrow$}    & (0.2,0.3{]} & 22.070 & 24.251 & 25.526  & 25.026          & 25.278          & 25.550 & \textbf{26.407} \\
                                                                                  &                                      & (0.3,0.4{]} & 20.066 & 22.249 & 23.347  & 23.166          & 23.076          & 23.271 & \textbf{24.007} \\
                                                                                  &                                      & (0.4,0.5{]} & 18.393 & 20.628 & 21.607  & 21.525          & 21.337          & 21.506 & \textbf{22.159} \\
                                                                                  &                                      & (0.5,0.6{]} & 16.139 & 18.550 & 19.270  & 19.278          & 19.009          & 19.127 & \textbf{19.622} \\ \cline{2-10} 
                                                                                  & \multirow{4}{*}{SSIM $^\uparrow$}    & (0.2,0.3{]} & 0.8388 & 0.8866 & 0.9076  & 0.8995          & 0.9033          & 0.9091 & \textbf{0.9225} \\
                                                                                  &                                      & (0.3,0.4{]} & 0.7701 & 0.8343 & 0.8602  & 0.8562          & 0.8536          & 0.8593 & \textbf{0.8776} \\
                                                                                  &                                      & (0.4,0.5{]} & 0.6940 & 0.7766 & 0.8072  & 0.8041          & 0.7982          & 0.8041 & \textbf{0.8266} \\
                                                                                  &                                      & (0.5,0.6{]} & 0.5936 & 0.6945 & 0.7247  & 0.7220          & 0.7096          & 0.7148 & \textbf{0.7398} \\ \cline{2-10} 
                                                                                  & \multirow{4}{*}{MAE $^\downarrow$}   & (0.2,0.3{]} & 0.0325 & 0.0216 & 0.0183  & 0.0185          & 0.0185          & 0.0185 & \textbf{0.0154} \\
                                                                                  &                                      & (0.3,0.4{]} & 0.0465 & 0.0315 & 0.0272  & 0.0269          & 0.0277          & 0.0277 & \textbf{0.0238} \\
                                                                                  &                                      & (0.4,0.5{]} & 0.0629 & 0.0429 & 0.0375  & 0.0369          & 0.0383          & 0.0383 & \textbf{0.0335} \\
                                                                                  &                                      & (0.5,0.6{]} & 0.0903 & 0.0617 & 0.0556  & 0.0546          & 0.0571          & 0.0572 & \textbf{0.0519} \\ \cline{2-10} 
                                                                                  & \multirow{4}{*}{LPIPS $^\downarrow$} & (0.2,0.3{]} & 0.1625 & 0.0941 & 0.0891  & 0.0982          & 0.0804          & 0.0922 & \textbf{0.0670} \\
                                                                                  &                                      & (0.3,0.4{]} & 0.2279 & 0.1369 & 0.1369  & 0.1269          & 0.1174          & 0.1443 & \textbf{0.1064} \\
                                                                                  &                                      & (0.4,0.5{]} & 0.3001 & 0.1850 & 0.1918  & 0.1665          & 0.1599          & 0.2022 & \textbf{0.1532} \\
                                                                                  &                                      & (0.5,0.6{]} & 0.3911 & 0.2601 & 0.2861  & \textbf{0.2336} & 0.2362          & 0.2974 & 0.2472          \\ \cline{2-10} 
                                                                                  & \multirow{4}{*}{FID $^\downarrow$}   & (0.2,0.3{]} & 13.477 & 5.0693 & 4.3334  & 4.2622          & 3.7142          & 5.2107 & \textbf{2.6902} \\
                                                                                  &                                      & (0.3,0.4{]} & 21.831 & 8.3719 & 7.9738  & 6.0240          & 5.8944          & 9.7656 & \textbf{4.7671} \\
                                                                                  &                                      & (0.4,0.5{]} & 33.158 & 13.247 & 13.580  & 9.0021          & 9.0499          & 16.902 & \textbf{7.9979} \\
                                                                                  &                                      & (0.5,0.6{]} & 47.170 & 21.758 & 26.355  & \textbf{15.400}          & 15.764          & 31.865 & 19.181          \\ \hline
\end{tabular}
\end{table*}
\ifx\alltabs\undefined
\end{document}
\fi

\section{Experiments}
\label{exp}
In this section, we compare our results with several SOTA inpainting methods, both quantitatively and qualitatively. Then, we conduct ablative experiments to study the effects of \ac{TPL} and \ac{TSL}, as well as the guidance metric and the hyperparameter K. In analysis, we study the superiority and the universality of our framework, and we show the weight adaptation processes.

\subsection{Comparisons with State-of-the-art Methods}

To demonstrate the effectiveness of our framework, we train CTSDG~\cite{guo2021ctsdg} with our framework, and compare our results with several state-of-the-art \ip methods. The models are: PIC~\cite{zheng2019pluralistic}(CVPR2019), EG~\cite{nazeri2019edgeconnect} (ICCV2019), RFR~\cite{li2020recurrent} (CVPR2020), DSNet~\cite{wang2021dynamic} (TIP2021) and CTSDG~\cite{guo2021ctsdg} (ICCV2021).

\emph{Qualitative Comparisons}. 
Fig.~\ref{fig:comp-methods} presents our results and results of existing \ip methods on CelebA, Paris StreetView and Places2. Pluralistic generates results with distorted structures. EG is limited in inferring object contours such as the jaw or the hood. RFR tends to produce artefacts like ink marks. The DSNet is also limited in producing determined structures, and CTSDG generates over blurred contents. Compared with these methods, our results obtain less structure ambiguity and texture inconsistency in the destroyed regions.

\begin{figure*}[t]
    \centering
    \centerline{\includegraphics[width=1\linewidth]{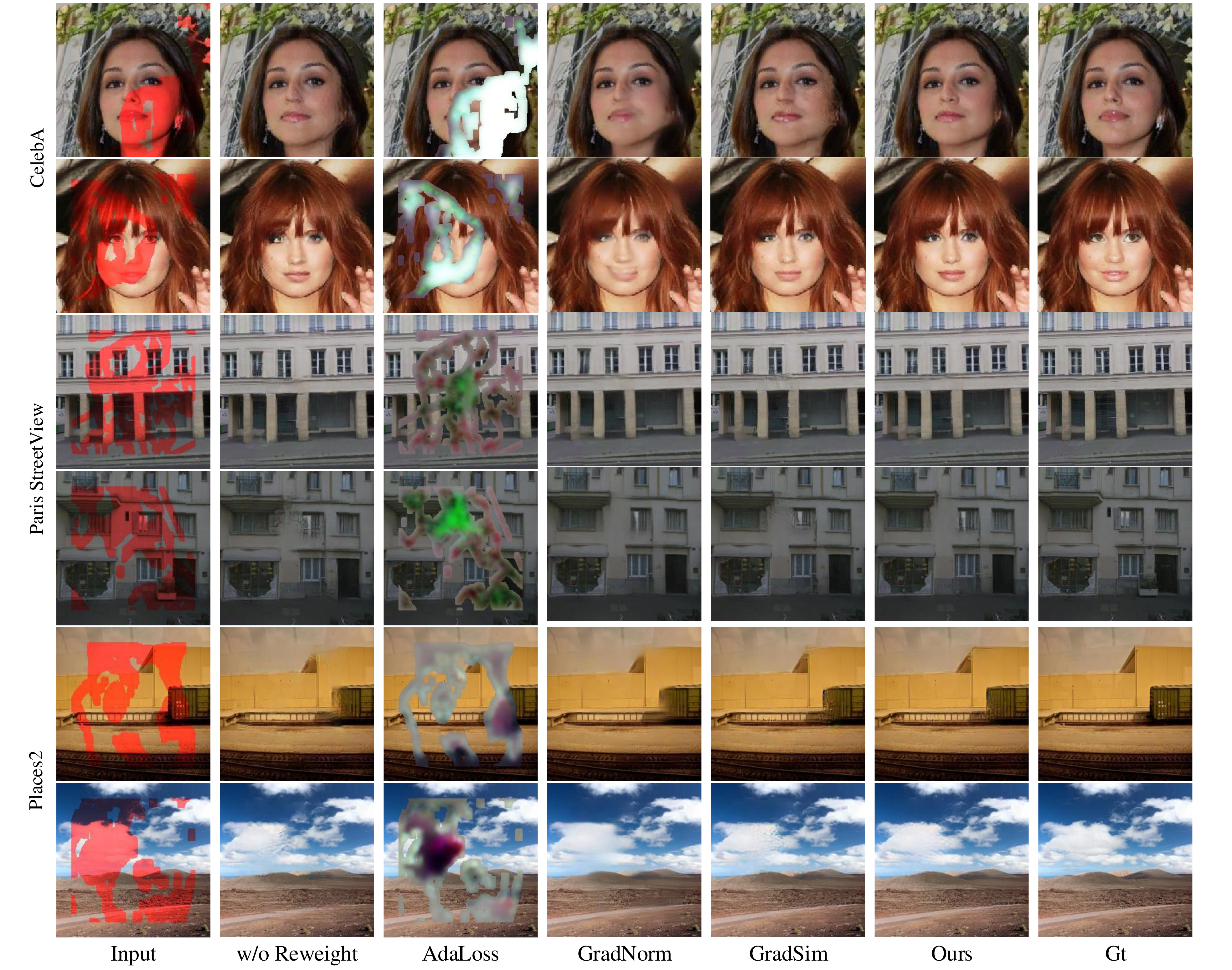}}
    \caption{Qualitative comparisons of our algorithm and other three loss reweighting algorithms. w/o Reweight denotes CTSDG trained with fixed auxiliary weights. The middle three lines are results of CTSDG trained with \ac{TPL} and \ac{TSL} reweighted by AdaLoss, GradNorm and GradSim respectively. Ours is CTSDG trained with our framework.}
    \label{fig:comp_algo}
\end{figure*}

\emph{Quantitative Comparisons}.
We also compare their results quantitatively. In this part, we evaluate \ip performances in terms of five commonly used metrics: PSNR, SSIM, MAE, LPIPS~\cite{zhang2018lpips} and FID~\cite{heusel2017fid}. Following \cite{zhang2018lpips,wang2021dynamic}, these metrics can be grouped into two classes of distortion metrics (PSNR, SSIM and MAE) and perception metrics (FID and LPIPS). Distortion metrics are used to quantify the distance between the ground truth image and generated image at the pixel level, while the perception metrics are introduced to measure the high level deep features of generated images. As we find the distortion metrics often contradict human judgements~\cite{zhang2018lpips}, we also use perception metrics to evaluate image qualities from a perspective close to humans. 

In TABLE~\ref{tab:effectiveness}, we summarize the performance of CTSDG trained by our framework and SOTA methods on datasets of Paris StreetView, CelebA and Places2. What stands out is that our results achieve the best distortion scores, compared with SOTA methods. In particular our framework consistently improves the performance of CTSDG by a large margin across all datasets. As for perception scores, we are better than other SOTA methods except RFR. Though our LPIPS scores are slightly weaker than RFR on CelebA and Paris StreetView, our results have better FID scores. These demonstrates that our framework can efficiently improve performance of current \ip method. 
\ifx\alltabs\undefined
\documentclass[lettersize,journal]{IEEEtran}
\usepackage{amsmath,amsfonts}
\usepackage{algorithmic}
\usepackage{algorithm}
\usepackage{array}
\usepackage[caption=false,font=normalsize,labelfont=sf,textfont=sf]{subfig}
\usepackage{textcomp}
\usepackage{stfloats}
\usepackage{url}
\usepackage{verbatim}
\usepackage{graphicx}
\usepackage{cite}
\hyphenation{op-tical net-works semi-conduc-tor IEEE-Xplore}

\begin{document}
\fi

\begin{table}[thbp]
\caption{Ablative experiments. The best results are $BoldFaced$. $^\uparrow$ Means Higher Is Better. $^\downarrow$ Means Lower Is better.\label{tab:ab1}}
\centering
\begin{tabular}{c|ccccc}
\hline
Experiment & PSNR $^\uparrow$ & SSIM $^\uparrow$ & MAE $^\downarrow$ & LPIPS $^\downarrow$ & FID $^\downarrow$ \\ \hline
1          & 32.084           & 0.9481           & 0.0128            & 0.0500              & 4.9752            \\
2          & 32.712           & 0.9535           & 0.0119            & 0.0435              & 5.1987            \\
3          & \textbf{32.909}  & 0.9541           & \textbf{0.0110}   & 0.0403              & \textbf{4.0748}   \\
4          & 32.799           & \textbf{0.9545}  & 0.0111            & 0.0425              & 4.5517            \\
5          & 32.905           & 0.9535           & \textbf{0.0110}   & \textbf{0.0402}     & 4.1045            \\
6          & 32.880           & 0.9541           & \textbf{0.0110}   & 0.0404              & 4.1787            \\ \hline
\end{tabular}
\end{table}
\ifx\alltabs\undefined
\end{document}
\fi
\subsection{Ablation Study}
\label{ablative}
In this section, we conduct ablative studies of \ac{TPL} and \ac{TSL} losses, as well as the guidance metric and the hyperparameter K of \ac{AWA}. We choose CTSDG as the base model and train generators on CelebA-HQ dataset. Specific settings of the six experiments are as follows:
\begin{itemize}
    \item Experiment 1: We train CTSDG with standard perceptual and style losses with static auxiliary weights.
    \item Experiment 2 : We train CTSDG with perceptual and style losses and apply \ac{AWA} to adapt auxiliary weights. The guidance metric is LPIPS and K is set to 1.
    \item Experiment 3: We train CTSDG with \ac{TPL} and \ac{TSL} and apply \ac{AWA} to adapt auxiliary weights. The guidance metric is LPIPS and K is set to 1.
    \item Experiment 4: We replace the guidance metric in Experiment 3 with MAE.
    \item Experiment 5: We set K in Experiment 3 to 5.
    \item Experiment 6: We set K in Experiment 3 to 10.
\end{itemize}

\ifx\alltabs\undefined
\documentclass[lettersize,journal]{IEEEtran}
\usepackage{amsmath,amsfonts}
\usepackage{algorithmic}
\usepackage{algorithm}
\usepackage{array}
\usepackage[caption=false,font=normalsize,labelfont=sf,textfont=sf]{subfig}
\usepackage{textcomp}
\usepackage{stfloats}
\usepackage{url}
\usepackage{verbatim}
\usepackage{graphicx}
\usepackage{cite}
\hyphenation{op-tical net-works semi-conduc-tor IEEE-Xplore}

\begin{document}
\fi

\begin{table*}[thbp]
\caption{Quantitative Comparisons of Our Algorithm and Other Three Loss Reweighting Algorithms. The Best Results Are \textbf{Boldfaced}. AL, GN and GS denote AdaLoss, GradNorm and GradSim respectively. $^\uparrow$ Means Higher Is Better. $^\downarrow$ Means Lower Is Better.\label{tab:exp1}}
\centering
\begin{tabular}{p{0.05\linewidth}|p{0.05\linewidth}|p{0.04\linewidth}p{0.05\linewidth}p{0.05\linewidth}p{0.04\linewidth}|p{0.04\linewidth}p{0.05\linewidth}p{0.05\linewidth}p{0.04\linewidth}|p{0.04\linewidth}p{0.05\linewidth}p{0.05\linewidth}p{0.04\linewidth}}
\hline
\multirow{2}{*}{Metrics}             & \multirow{2}{*}{Mask} & \multicolumn{4}{c}{Paris StreetView}                & \multicolumn{4}{c}{CelebA}                 & \multicolumn{4}{c}{Places2}                         \\ \cline{3-14} 
                                     &                       & AL     & GN              & GS     & Ours            & AL     & GN     & GS     & Ours            & AL     & GN              & GS     & Ours            \\ \hline
\multirow{5}{*}{PSNR $^\uparrow$}    & (0.1,0.2{]}           & 23.132 & \textbf{33.730} & 33.146 & 33.520          & 19.695 & 32.988 & 32.809 & \textbf{33.073} & 23.208 & \textbf{30.229} & 29.685 & 29.971          \\
                                     & (0.2,0.3{]}           & 20.352 & \textbf{30.272} & 29.639 & 29.942          & 15.925 & 29.458 & 29.346 & \textbf{29.610} & 20.432 & \textbf{26.731} & 26.154 & 26.407          \\
                                     & (0.3,0.4{]}           & 18.707 & \textbf{27.855} & 27.258 & 27.664          & 12.373 & 26.970 & 26.937 & \textbf{27.202} & 18.448 & \textbf{24.349} & 23.772 & 24.007          \\
                                     & (0.4,0.5{]}           & 16.593 & \textbf{25.792} & 25.196 & 25.554          & 9.417  & 24.982 & 24.997 & \textbf{25.261} & 16.734 & \textbf{22.498} & 21.935 & 22.159          \\
                                     & (0.5,0.6{]}           & 12.538 & \textbf{22.687} & 22.339 & 22.502          & 7.025  & 21.898 & 22.057 & \textbf{22.294} & 13.874 & \textbf{19.903} & 19.422 & 19.622          \\ \hline
\multirow{5}{*}{SSIM $^\uparrow$}    & (0.1,0.2{]}           & 0.8665 & \textbf{0.9710} & 0.9678 & 0.9695          & 0.8431 & 0.9749 & 0.9739 & \textbf{0.9752} & 0.8869 & \textbf{0.9616} & 0.9584 & 0.9605          \\
                                     & (0.2,0.3{]}           & 0.7879 & \textbf{0.9452} & 0.9391 & 0.9418          & 0.7534 & 0.9507 & 0.9492 & \textbf{0.9518} & 0.8166 & \textbf{0.9252} & 0.9187 & 0.9225          \\
                                     & (0.3,0.4{]}           & 0.7242 & \textbf{0.9108} & 0.9014 & 0.9059          & 0.6664 & 0.9217 & 0.9200 & \textbf{0.9240} & 0.7497 & \textbf{0.8823} & 0.8720 & 0.8776          \\
                                     & (0.4,0.5{]}           & 0.6475 & \textbf{0.8690} & 0.8549 & 0.8609          & 0.5747 & 0.8885 & 0.8865 & \textbf{0.8921} & 0.6793 & \textbf{0.8337} & 0.8190 & 0.8266          \\
                                     & (0.5,0.6{]}           & 0.5616 & \textbf{0.7941} & 0.7723 & 0.7807          & 0.5150 & 0.8260 & 0.8251 & \textbf{0.8329} & 0.5870 & \textbf{0.7526} & 0.7296 & 0.7398          \\ \hline
\multirow{5}{*}{MAE $^\downarrow$}   & (0.1,0.2{]}           & 0.0244 & \textbf{0.0065} & 0.0069 & 0.0067          & 0.0362 & 0.0054 & 0.0055 & \textbf{0.0053} & 0.0223 & \textbf{0.0079} & 0.0084 & 0.0081          \\
                                     & (0.2,0.3{]}           & 0.0417 & \textbf{0.0111} & 0.0118 & 0.0115          & 0.0724 & 0.0103 & 0.0104 & \textbf{0.0100} & 0.0391 & \textbf{0.0149} & 0.0159 & 0.0154          \\
                                     & (0.3,0.4{]}           & 0.0578 & \textbf{0.0168} & 0.0179 & 0.0173          & 0.1277 & 0.0160 & 0.0161 & \textbf{0.0155} & 0.0572 & \textbf{0.0231} & 0.0246 & 0.0238          \\
                                     & (0.4,0.5{]}           & 0.0817 & \textbf{0.0240} & 0.0256 & 0.0248          & 0.2037 & 0.0228 & 0.0227 & \textbf{0.0220} & 0.0784 & \textbf{0.0326} & 0.0346 & 0.0335          \\
                                     & (0.5,0.6{]}           & 0.1439 & \textbf{0.0388} & 0.0403 & 0.0398          & 0.3013 & 0.0375 & 0.0366 & \textbf{0.0354} & 0.1216 & \textbf{0.0509} & 0.0534 & 0.0519          \\ \hline
\multirow{5}{*}{LPIPS $^\downarrow$} & (0.1,0.2{]}           & 0.2167 & 0.0384          & 0.0344 & \textbf{0.0338} & 0.2332 & 0.0351 & 0.0270 & \textbf{0.0275} & 0.1655 & 0.0440          & 0.0388 & \textbf{0.0345} \\
                                     & (0.2,0.3{]}           & 0.3134 & 0.0717          & 0.0645 & \textbf{0.0630} & 0.3355 & 0.0686 & 0.0513 & \textbf{0.0521} & 0.2567 & 0.0854          & 0.0753 & \textbf{0.0670} \\
                                     & (0.3,0.4{]}           & 0.3894 & 0.1176          & 0.1036 & \textbf{0.1014} & 0.4299 & 0.1081 & 0.0794 & \textbf{0.0800} & 0.3407 & 0.1360          & 0.1195 & \textbf{0.1064} \\
                                     & (0.4,0.5{]}           & 0.4760 & 0.1744          & 0.1532 & \textbf{0.1529} & 0.5286 & 0.1526 & 0.1116 & \textbf{0.1112} & 0.4238 & 0.1941          & 0.1711 & \textbf{0.1532} \\
                                     & (0.5,0.6{]}           & 0.6018 & 0.2845          & 0.2440 & \textbf{0.2408} & 0.6160 & 0.2364 & 0.1721 & \textbf{0.1675} & 0.5288 & 0.3081          & 0.2677 & \textbf{0.2472} \\ \hline
\multirow{5}{*}{FID $^\downarrow$}   & (0.1,0.2{]}           & 188.12 & 25.518          & 23.160 & \textbf{20.545} & 89.539 & 1.3182 & 1.2559 & \textbf{1.1591} & 33.072 & 1.9996          & 1.4938 & \textbf{1.2899} \\
                                     & (0.2,0.3{]}           & 239.00 & 39.069          & 35.356 & \textbf{30.985} & 131.42 & 3.4296 & 3.0543 & \textbf{2.7465} & 61.879 & 5.0628          & 3.2156 & \textbf{2.6902} \\
                                     & (0.3,0.4{]}           & 244.91 & 56.512          & 46.829 & \textbf{39.842} & 169.32 & 7.1881 & 5.9509 & \textbf{5.2663} & 86.089 & 10.659          & 5.9366 & \textbf{4.7671} \\
                                     & (0.4,0.5{]}           & 265.12 & 80.858          & 65.225 & \textbf{53.016} & 206.56 & 13.096 & 10.068 & \textbf{8.8076} & 107.32 & 19.789          & 10.360 & \textbf{7.9979} \\
                                     & (0.5,0.6{]}           & 286.96 & 121.88          & 92.273 & \textbf{76.936} & 223.02 & 25.623 & 18.147 & \textbf{16.396} & 128.04 & 42.606          & 21.450 & \textbf{19.181} \\ \hline
\end{tabular}
\end{table*}
\ifx\alltabs\undefined
\end{document}
\fi
TABLE~\ref{tab:ab1} shows the results of these experiments. Comparing Experiment 1 with others, it is obvious that the \ip performance is improved by a large margin by our framework. Compared with Experiment 2, the inpainting performance of Experiment 3 is consistently increased, which demonstrates the superiority of the proposed \ac{TPL} and \ac{TSL}. Though the distortion scores of Experiment 3 are close to Experiment 4, it has better perceptual scores, which reveals that LPIPS is more fit for guiding auxiliary weights in the \ac{AWA} algorithm. Comparing the results of Experiments 3, 5 and 6, we conclude that the \ac{AWA} algorithm is not sensitive to the hyperparameter K, as they have similar \ip performance.

\subsection{Analysis}
\emph{Superiority of AWA}.  We select CTSDG as the base model, and apply other three loss reweighting algorithms to adjust weights of \ac{TPL} and \ac{TSL}.
\begin{itemize}
    \item \emph{GradSim}\cite{du2018gradsim}: GradSim reweights auxiliary losses based on the similarity between auxiliary losses and main loss. In effect, an auxiliary loss is manually filtered out when it has an opposite gradient direction to the main loss.
    \item \emph{GradNorm}\cite{chen2018gradnorm}: GradNorm is a algorithm that automatically reweights loses using trainable parameters. In GradNorm, weights are tuned to balance the gradient norms among losses.
    \item \emph{AdaLoss}\cite{hu2019adaloss}: AdaLoss is a principled algorithm that makes each weight inversely proportional to the empirical mean of the corresponding loss on the training set.
\end{itemize}

As Fig.~\ref{fig:comp_algo} shows, reweighted by AdaLoss, generators fail to converge. Generators tend to produce blurry contents when using GradNorm. GradSim still struggles to guide models generating determined contents such as noses and windows. In contrast, with \ac{AWA}, auxiliary weights help generate contents with consistent textures and determined structures.

We also compare their results quantitatively. In in TABLE~\ref{tab:exp1}, AL, GN and GS denote AdaLoss, GradNorm and GradSim respectively. As TABLE~\ref{tab:exp1} shows, our results achieve the best performances in terms of perception metrics across all datasets which is consistent with qualitative results. As for distortion metrics, our algorithm achieves the best on CelebA, but achieves near best on the other datasets. We conjecture the reason is that the GradNorm (GN) achieves the lowest distortion scores by averaging possible patterns and generating blurry contents (see~\Figref{fig:comp_algo}).

\begin{figure*}[htbp]
    \centering
    \centerline{\includegraphics[width=1.\linewidth]{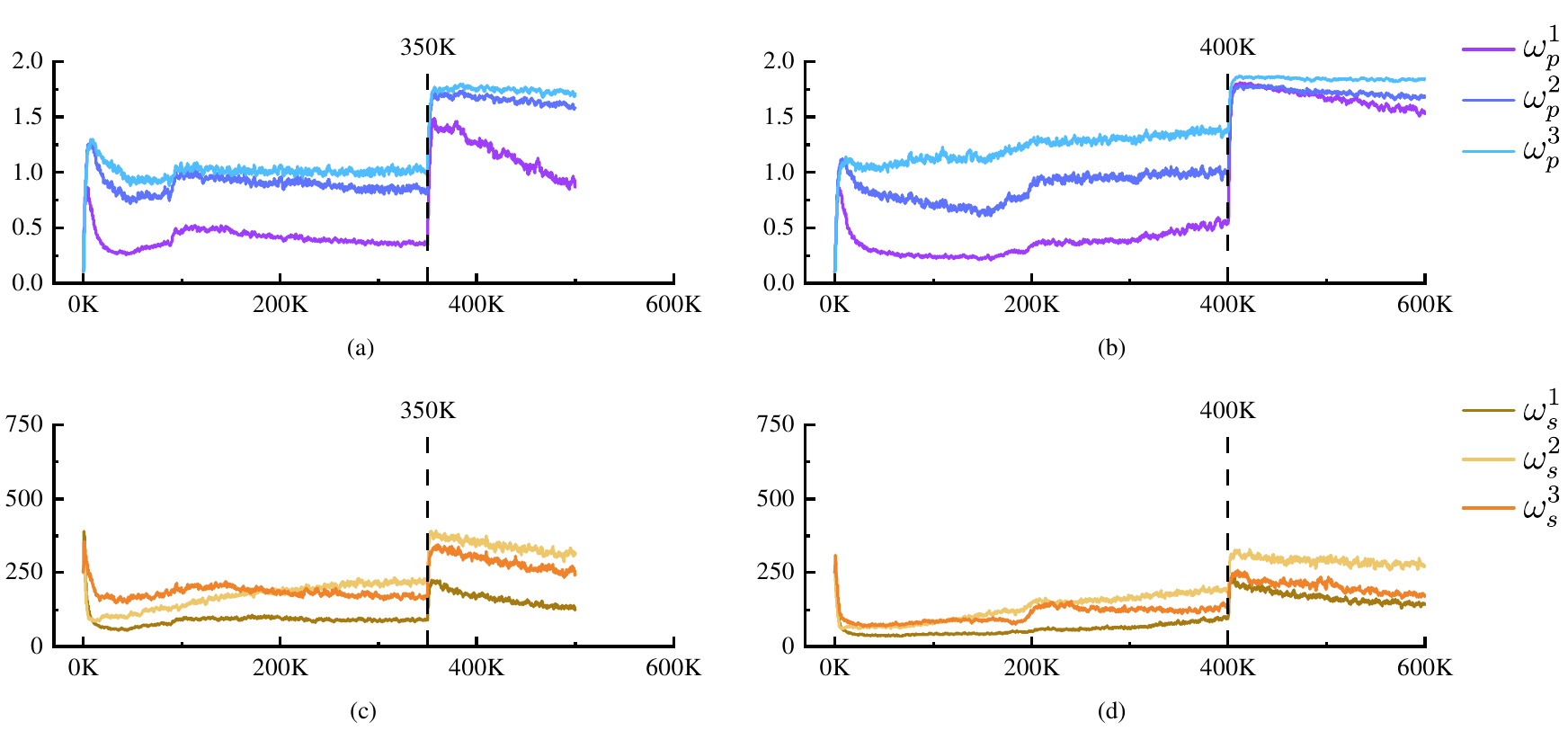}}
    \caption{Adaptation processes of \ac{TPL} and \ac{TSL}. (a) and (c) are weight adaptations of TPL and TSL on the CelebA dataset, where $\omega_p^1$, $\omega_p^2$ and $\omega_p^3$ are weights of \ac{TPL} corresponding to the first, second and third max-pooling layers of the VGG-16 network respectively. (b) and (d) show how weights of TPL and TSL adapt on Paris StreetView dataset. $\omega_s^1$, $\omega_s^2$ and $\omega_s^3$ denote the weights of \ac{TSL}, and they also correspond to the first three max-pooling layers of the VGG-16 network, in shallow to deep manner.}
    \label{fig:adaw}
\end{figure*}

\ifx\alltabs\undefined
\documentclass[lettersize,journal]{IEEEtran}
\usepackage{amsmath,amsfonts}
\usepackage{algorithmic}
\usepackage{algorithm}
\usepackage{array}
\usepackage[caption=false,font=normalsize,labelfont=sf,textfont=sf]{subfig}
\usepackage{textcomp}
\usepackage{stfloats}
\usepackage{url}
\usepackage{verbatim}
\usepackage{graphicx}
\usepackage{cite}
\hyphenation{op-tical net-works semi-conduc-tor IEEE-Xplore}

\begin{document}
\fi

\begin{table}[thbp]
\caption{Quantitative Results Of two base models trained by customary settings and our framework on CelebA dataset. The Best Results Are \textbf{Boldfaced}. $^\uparrow$ Means Higher Is Better. Ours means the model is trained by our framework. $^\downarrow$ Means Lower Is Better.\label{tab:generality1}}
\centering
\begin{tabular}{p{0.11\linewidth}|p{0.11\linewidth}|p{0.11\linewidth}p{0.11\linewidth}|p{0.11\linewidth}p{0.11\linewidth}}
\hline
Metric                               & Mask        & PRVS            & Ours             & EG2             & Ours             \\ \hline
\multirow{5}{*}{PSNR $^\uparrow$}    & (0.1,0.2{]} & 31.860          & \textbf{32.191}  & 33.747          & \textbf{34.072}  \\
                                     & (0.2,0.3{]} & 28.633          & \textbf{28.989}  & 31.164          & \textbf{31.477}  \\
                                     & (0.3,0.4{]} & 26.373          & \textbf{26.737}  & 29.352          & \textbf{29.669}  \\
                                     & (0.4,0.5{]} & 24.525          & \textbf{24.886}  & 27.860          & \textbf{28.186}  \\
                                     & (0.5,0.6{]} & 21.750          & \textbf{22.045}  & 25.611          & \textbf{25.879}  \\ \hline
\multirow{5}{*}{SSIM $^\uparrow$}    & (0.1,0.2{]} & 0.9684          & \textbf{0.9694}  & 0.9758          & \textbf{0.9775}  \\
                                     & (0.2,0.3{]} & 0.9417          & \textbf{0.9435}  & 0.9587          & \textbf{0.9616}  \\
                                     & (0.3,0.4{]} & 0.9113          & \textbf{0.9141}  & 0.9405          & \textbf{0.9449}  \\
                                     & (0.4,0.5{]} & 0.8767          & \textbf{0.8807}  & 0.9207          & \textbf{0.9271}  \\
                                     & (0.5,0.6{]} & 0.8148          & \textbf{0.8210}  & 0.8890          & \textbf{0.8985}  \\ \hline
\multirow{5}{*}{MAE $^\downarrow$}   & (0.1,0.2{]} & 0.0066          & \textbf{0.0065}  & 0.0059          & \textbf{0.0057}  \\
                                     & (0.2,0.3{]} & 0.0120          & \textbf{0.0118}  & 0.0100          & \textbf{0.0096}  \\
                                     & (0.3,0.4{]} & 0.0179          & \textbf{0.0178}  & 0.0142          & \textbf{0.0137}  \\
                                     & (0.4,0.5{]} & \textbf{0.0249} & \textbf{0.0249}  & 0.0189          & \textbf{0.0183}  \\
                                     & (0.5,0.6{]} & 0.0394          & \textbf{0.0388}  & 0.0273          & \textbf{0.0266}  \\ \hline
\multirow{5}{*}{LPIPS $^\downarrow$} & (0.1,0.2{]} & 0.0362          & \textbf{0.0305}  & \textbf{0.0231} & \textbf{0.0231}  \\
                                     & (0.2,0.3{]} & 0.0678          & \textbf{0.0558}  & 0.0401          & \textbf{0.0397}  \\
                                     & (0.3,0.4{]} & 0.1034          & \textbf{0.0842}  & 0.0585          & \textbf{0.0575}  \\
                                     & (0.4,0.5{]} & 0.1435          & \textbf{0.1164}  & 0.0787          & \textbf{0.0771}  \\
                                     & (0.5,0.6{]} & 0.2138          & \textbf{0.1720}  & 0.1092          & \textbf{0.1070}  \\ \hline
\multirow{5}{*}{FID $^\downarrow$}   & (0.1,0.2{]} & 1.7208          & \textbf{1.4015}  & 1.0978          & \textbf{1.0225}  \\
                                     & (0.2,0.3{]} & 4.4684          & \textbf{3.4639}  & 2.3558          & \textbf{2.2207}  \\
                                     & (0.3,0.4{]} & 8.7892          & \textbf{6.7080}  & 4.0951          & \textbf{3.8143}  \\
                                     & (0.4,0.5{]} & 15.341         & \textbf{11.692} & 6.5822          & \textbf{6.1479}  \\
                                     & (0.5,0.6{]} & 27.478         & \textbf{23.910} & 10.725         & \textbf{10.017} \\ \hline
\end{tabular}
\end{table}

\ifx\alltabs\undefined
\end{document}
\fi
\ifx\alltabs\undefined
\documentclass[lettersize,journal]{IEEEtran}
\usepackage{amsmath,amsfonts}
\usepackage{algorithmic}
\usepackage{algorithm}
\usepackage{array}
\usepackage[caption=false,font=normalsize,labelfont=sf,textfont=sf]{subfig}
\usepackage{textcomp}
\usepackage{stfloats}
\usepackage{url}
\usepackage{verbatim}
\usepackage{graphicx}
\usepackage{cite}
\hyphenation{op-tical net-works semi-conduc-tor IEEE-Xplore}

\begin{document}
\fi

\begin{table}[thbp]
\caption{Quantitative Results Of two base models trained by customary settings and our framework on Paris StreetView dataset. Ours means the model is trained with our framework. The Best Results Are \textbf{Boldfaced}. $^\uparrow$ Means Higher Is Better. $^\downarrow$ Means Lower Is Better.\label{tab:generality2}}
\centering
\begin{tabular}{p{0.11\linewidth}|p{0.11\linewidth}|p{0.11\linewidth}p{0.11\linewidth}|p{0.11\linewidth}p{0.11\linewidth}}
\hline
\multicolumn{1}{l}{Metric}           & \multicolumn{1}{l|}{Mask} & PRVS            & Ours            & EG2             & Ours            \\ \hline
\multirow{5}{*}{PSNR $^\uparrow$}    & (0.1,0.2{]}               & 31.380          & \textbf{31.391} & 33.282          & \textbf{33.417} \\
                                     & (0.2,0.3{]}               & 28.354          & \textbf{28.451} & 30.895          & \textbf{30.992} \\
                                     & (0.3,0.4{]}               & 26.241          & \textbf{26.488} & 29.151          & \textbf{29.271} \\
                                     & (0.4,0.5{]}               & 24.126          & \textbf{24.690} & 27.547          & \textbf{27.692} \\
                                     & (0.5,0.6{]}               & 21.419          & \textbf{22.214} & 25.159          & \textbf{25.201} \\ \hline
\multirow{5}{*}{SSIM $^\uparrow$}    & (0.1,0.2{]}               & 0.9565          & \textbf{0.9565} & 0.9694          & \textbf{0.9701} \\
                                     & (0.2,0.3{]}               & 0.9233          & \textbf{0.9242} & 0.9489          & \textbf{0.9504} \\
                                     & (0.3,0.4{]}               & 0.8835          & \textbf{0.8867} & 0.9260          & \textbf{0.9285} \\
                                     & (0.4,0.5{]}               & 0.8315          & \textbf{0.8395} & 0.8990          & \textbf{0.9028} \\
                                     & (0.5,0.6{]}               & 0.7446          & \textbf{0.7647} & 0.8517          & \textbf{0.8556} \\ \hline
\multirow{5}{*}{MAE $^\downarrow$}   & (0.1,0.2{]}               & 0.0120          & \textbf{0.0120} & 0.0106          & \textbf{0.0105} \\
                                     & (0.2,0.3{]}               & 0.0173          & \textbf{0.0171} & 0.0144          & \textbf{0.0143} \\
                                     & (0.3,0.4{]}               & 0.0235          & \textbf{0.0229} & 0.0184          & \textbf{0.0182} \\
                                     & (0.4,0.5{]}               & 0.0320          & \textbf{0.0304} & 0.0236          & \textbf{0.0232} \\
                                     & (0.5,0.6{]}               & 0.0477          & \textbf{0.0432} & 0.0328          & \textbf{0.0326} \\ \hline
\multirow{5}{*}{LPIPS $^\downarrow$} & (0.1,0.2{]}               & 0.0881          & \textbf{0.0490} & 0.0367          & \textbf{0.0364} \\
                                     & (0.2,0.3{]}               & 0.1133          & \textbf{0.0804} & 0.0545          & \textbf{0.0539} \\
                                     & (0.3,0.4{]}               & 0.1455          & \textbf{0.1173} & 0.0749          & \textbf{0.0732} \\
                                     & (0.4,0.5{]}               & 0.1931          & \textbf{0.1666} & 0.1006          & \textbf{0.0988} \\
                                     & (0.5,0.6{]}               & 0.2750          & \textbf{0.2445} & 0.1396          & \textbf{0.1373} \\ \hline
\multirow{5}{*}{FID $^\downarrow$}   & (0.1,0.2{]}               & \textbf{22.778} & 23.682          & 18.941          & \textbf{18.528} \\
                                     & (0.2,0.3{]}               & \textbf{35.892} & 36.941          & 27.415          & \textbf{27.301} \\
                                     & (0.3,0.4{]}               & 48.371          & \textbf{47.383} & 36.976          & \textbf{35.401} \\
                                     & (0.4,0.5{]}               & 68.103          & \textbf{65.068} & 45.613          & \textbf{45.466} \\
                                     & (0.5,0.6{]}               & 99.863          & \textbf{87.485} & \textbf{58.323} & 58.767          \\ \hline
\end{tabular}
\end{table}

\ifx\alltabs\undefined
\end{document}
\fi

\emph{Universality of AWA}. We also apply our framework to train various SOTA methods~\cite{nazeri2019edgeconnect,li2019progressive} on Paris StreetView and CelebA datasets. As EG has two generators and only the image generator (EG2) is trained by perceptual and style losses, we choose EG2 for experiments. As presented in~\Tabref{tab:generality1} and~\Tabref{tab:generality2}, AWA can consistently improve the inpainting performance of SOTA methods. This indicates that \ac{AWA} is not limited to one specific inpainting architecture but can be easily generalized to help improve the ability of auxiliary losses for image inpainting. 

\emph{Weight Adaptation Processes}.
It is also insightful to observe how auxiliary weights adapt during training processes. Models here are CTSDG trained by our framework. Fig.~\ref{fig:adaw} show the adaption processes of \ac{TPL} and \ac{TSL} on CelebA and Paris StreetView datasets. It is obvious that all weights increase at the start of fine-tune stage (350,000 for CelebA and 400,000 for Paris StreetView), which reveals that deep feature based auxiliary losses are more useful in the fine-tune stage. As Fig.~\ref{fig:adaw}(a) and Fig.~\ref{fig:adaw} (b) show, $\omega_p^1$ is the smallest while $\omega_p^3$ is the largest. This indicates that loss terms of shallower features are paid less attention. The similar phenomena can also be found in the adaptation processes of \ac{TSL} (Fig.~\ref{fig:adaw}(c) and Fig.~\ref{fig:adaw}(d)). We conjecture that deeper features are more useful for \ac{TPL} and \ac{TSL} to train inpainting generators. 

\ifx\allfiles\undefined
\bibliographystyle{IEEEtran}
\bibliography{reference/reference.bib}
\end{document}
\fi
\ifx\allfiles\undefined
\documentclass[lettersize,journal]{IEEEtran}
\usepackage{amsmath,amsfonts}
\usepackage{algorithm}
\usepackage{array}
\usepackage[caption=false,font=normalsize,labelfont=sf,textfont=sf]{subfig}
\usepackage{textcomp}
\usepackage{stfloats}
\usepackage{url}
\usepackage{verbatim}
\usepackage{graphicx}
\usepackage{cite}
\hyphenation{op-tical net-works semi-conduc-tor IEEE-Xplore}

\begin{document}
\linenumbers
\fi

\section{Discussions and Future Work}
In this paper, we propose an auxiliary loss reweighting framework for image inpainting, which train inpainting generators with better auxiliary losses and efficiently reweight auxiliary losses without time-consuming grid search. Specifically, we propose \ac{TPL} and \ac{TSL} to increase the effectiveness of standard perceptual loss and style loss by independently weighting different loss terms of different deep features. We further design the \ac{AWA} algorithm to dynamically adjusts the parameters of \ac{TPL} and \ac{TSL} in a single training process. Our framework is effective and generalizes well for the existing image inpainting models.

Though the proposed framework is efficient in image inpainting, the magnitude of auxiliary parameters is restricted by the requirement of computing Jacobian matrices in the auxiliary update process. In effect, the computational cost linearly increases as the number of auxiliary parameters grows.

We believe that our approach is one step toward a simple and general auxiliary loss adaptation framework that could dynamically reweight auxiliary losses for inpainting models with better performance and less computational cost. Extending such capacity to other tasks and increasing the flexibility of our method is ongoing.

\ifx\allfiles\undefined
\bibliographystyle{IEEEtran}
\bibliography{reference/reference.bib,reference/diffusion_based.bib,reference/patch_based.bib,reference/diffusion_patch_based.bib,reference/loss_reweighting_alg.bib}
\end{document}
\fi
\ifx\allfiles\undefined
\documentclass[lettersize,journal]{IEEEtran}
\usepackage{amsmath,amsfonts}
\usepackage{algorithm}
\usepackage{array}
\usepackage[caption=false,font=normalsize,labelfont=sf,textfont=sf]{subfig}
\usepackage{textcomp}
\usepackage{stfloats}
\usepackage{url}
\usepackage{verbatim}
\usepackage{graphicx}
\usepackage{cite}
\hyphenation{op-tical net-works semi-conduc-tor IEEE-Xplore}

\begin{document}
\linenumbers
\fi

\ifx\allfiles\undefined
\bibliographystyle{IEEEtran}
\bibliography{reference/reference.bib,reference/diffusion_based.bib,reference/patch_based.bib,reference/diffusion_patch_based.bib,reference/loss_reweighting_alg.bib}
\end{document}
\fi
\ifx\allfiles\undefined
\documentclass[lettersize,journal]{IEEEtran}
\usepackage{amsmath,amsfonts}
\usepackage{algorithm}
\usepackage{array}
\usepackage[caption=false,font=normalsize,labelfont=sf,textfont=sf]{subfig}
\usepackage{textcomp}
\usepackage{stfloats}
\usepackage{url}
\usepackage{verbatim}
\usepackage{graphicx}
\usepackage{cite}
\hyphenation{op-tical net-works semi-conduc-tor IEEE-Xplore}


\begin{document}
\linenumbers
\fi

\renewcommand{\theequation}{A\arabic{equation}}
\setcounter{equation}{0}

\def \philoss {\is(\The^K)}

\appendix
Here we prove that the surrogate loss function equals $\is$ by verifying that they share the same gradient directions with respect to auxiliary parameters of \ac{TPL} and \ac{TSL}. Recall that the original objective for auxiliary parameters is shown in~\Eqref{phiobj}. 
We first compute the gradient of $\is$ in~\Eqref{phiobj} with respect to auxiliary paramters. More formally, compute $\nabla_{\phi_p}\philoss$ and $\nabla_{\phi_s}\philoss$. Based on chain rules:
\begin{equation}
\label{a1}
\nabla_{\phi_p}\philoss=\note{\nabla_{\The^K}\is(\The^K)}{1\times|\The|}\note{J_{\The^K}(\omega_p)}{|\The|\times|\omega_p|}\note{J_{\omega_p}(\phi_p)}{|\omega_p|\times|\phi_p|},
\end{equation}
\begin{equation}
\label{a2}
  \nabla_{\phi_s}\philoss=\note{\nabla_{\The^K}\is(\The^K)}{1\times|\The|}\note{J_{\The^K}(\omega_s)}{|\The|\times|\omega_s|}\note{J_{\omega_s}(\phi_s)}{|\omega_s|\times|\phi_s|}.
\end{equation}

Next, we compute Jacobian matrices of $\The^K$ with respect to auxiliary weights $\omega_p$ and $\omega_s$. Following the definition of the Jacobian matrix, the element at location $(a,b)$ of $J_{\The^K}(\omega_p)$ denoted as $J_{\The^K}(\omega_p)_{(a,b)}$, which can be computed as follows:
\begin{equation}
\begin{aligned}
&J_{\The^K}(\omega_p)_{(a,b)}=\frac{\partial (\The^K)_a}{\partial(\omega_p)_b}\\
&=\frac{\partial \left( \theta -\alpha \sum_{j=1}^K{\left( \lambda _{m}^{T}J_m\left( \theta ^j \right) +\omega _{p}^{T}J_p\left( \theta ^j \right) +\omega _{s}^{T}J_s\left( \theta ^j \right) \right)} \right) _a}{\partial \left( \omega _p \right) _b}\\
&=-\alpha \sum_{j=1}^K \frac{\partial \left({\omega _{p}^{T}J_p\left( \theta ^j \right)} \right)_a}{\partial \left( \omega _p \right) _b}\\
&=-\alpha \sum_{j=1}^K J_p(\The^j)_{(b,a)}.
\end{aligned}
\end{equation}
The elements of $J_{\The^K}(\omega_p)$ shares the same computational process:
\begin{equation}
\begin{aligned}
&J_{\The^K}(\omega_s)_{(a,b)}=\frac{\partial (\The^K)_a}{\partial(\omega_s)_b}\\
&=\frac{\partial \left( \theta -\alpha \sum_{j=1}^K{\left( \lambda _{m}^{T}J_m\left( \theta ^j \right) +\omega _{p}^{T}J_p\left( \theta ^j \right) +\omega _{s}^{T}J_s\left( \theta ^j \right) \right)} \right) _a}{\partial \left( \omega _s \right) _b}\\
&=-\alpha \sum_{j=1}^K \frac{\partial \left({\omega_{s}^{T}J_s\left( \theta ^j \right)} \right)_a}{\partial \left( \omega_s \right) _b}\\
&=-\alpha \sum_{j=1}^K J_s(\The^j)_{(b,a)}.
\end{aligned}
\end{equation}
Such that these Jacobian matrices are:
\begin{equation}
\label{a5}
    J_{\The^K}(\omega_p)=-\alpha \sum_{j=1}^K J_p^T(\The^j),
\end{equation}
\begin{equation}
\label{a6}
J_{\The^K}(\omega_s)=-\alpha \sum_{j=1}^K J_s^T(\The^j).
\end{equation}


Then, combining~\Eqref{a1},~\Eqref{a5}, we get the explicit gradient of $\is$ with respect to auxiliary parameters of \ac{TPL}.
\begin{equation}
\label{a7}
\begin{aligned}
\nabla_{\phi_p}\philoss=&\nabla_{\The^K}\is(\The^K)J_{\The^K}(\omega_p)J_{\omega_p}(\phi_p)\\
&=-\alpha\nabla_{\The^K}\is(\The^K)\sum_{j=1}^K J_p^T(\The^j)J_{\omega_p}(\phi_p),
\end{aligned}
\end{equation}

\begin{equation}
\label{a8}
\begin{aligned}
\nabla_{\phi_s}\philoss=&\nabla_{\The^K}\is(\The^K)J_{\The^K}(\omega_s)J_{\omega_s}(\phi_s)\\
&=-\alpha\nabla_{\The^K}\is(\The^K)\sum_{j=1}^K J_s^T(\The^j)J_{\omega_s}(\phi_s).
\end{aligned}
\end{equation}

Finally, let's verify that the gradients of the surrogate loss in~\Eqref{eq:lossaux} share the same directions of the gradients of $\is$.
\begin{equation}
\begin{aligned}
\nabla_{\phi_p}\philoss&\propto\de\pp \L_{sg}(\phi_p,\phi_s)\\
&=-\nabla_\The\L_c(\The^K) \sum_j J_p(\The^j)J_{\omega_p}(\phi_p),
\end{aligned}
\end{equation}
\begin{equation}
\begin{aligned}
\nabla_{\phi_s}\philoss&\propto\de\pp \L_{sg}(\phi_s,\phi_s)\\
&=-\nabla_\The\L_c(\The^K) \sum_j J_s(\The^j)J_{\omega_s}(\phi_s),
\end{aligned}
\end{equation}

\ifx\allfiles\undefined
\end{document}
\fi

\bibliographystyle{IEEEtran}


\end{document}


\linenumbers
\fi

\newcommand{\de}[1] {{\nabla_{#1}}}
\def \domega {\nabla_{\omega}}
\def \dtheta {\nabla_{\theta}}
\def \is {\mathcal{L}_{s}}
\def \ic {\mathcal{L}_{t}}
\def \sgloss {\L_{sg}(\phi_p,\phi_s)}
\appendix
Here we illustrate the details of the derivation of~\Eqref{eq:lossaux}. Recall that the objective of auxiliary parameters of \ac{TPL} and \ac{TSL} is shown in~\Eqref{phiobj}. We firstly compute the explicit form of gradients of $\L_{sg}(\phi_p,)$,

However, it is impractical to directly to update auxiliary weights with Eq.~\eqref{eq:auxgrad}. As the objective for auxiliary weights in Eq.~\eqref{object2} is computed using the updated parameters of model parameters $\uk$, computing the gradients of auxiliary weights need to differentiate through the update process of model parameters, which  useless computational and memory cost to .as the differentiation through  would bring computational burdens.

Note that  when $U$ corresponds to gradient descent, $\uk$ represents adding a sequence of gradient vectors ($g_i$) to the initial model parameters $\theta$,
\begin{align}
\label{eq:uk}
 \uk&=\theta-\alpha\sum_{i=1}^k(g_i) \\ \label{eq:auxgrad2}
 g_i&=g_i^m+\lambda^T g_i^a
\end{align}
where $\alpha$ is the learning rate, $g_i^m$ and $g_i^a$ are the Jacobian matrix of main and auxiliary losses at step i. With Eq.~\eqref{eq:auxgrad2} we could eliminate unnecessary computational and memory cost, and rewrite the partial derivatives of auxiliary weights respect to $\uk$ as follows:
\begin{equation}
\label{eq:auxgrad2}
    \begin{aligned}
          \dlambda\uk=\sum_{i=1}^k (g_i^a)^T
    \end{aligned}
\end{equation}

When $U$ corresponds to Adam, the partial derivatives is:
\begin{equation}
\label{eq:auxgrad3}
\begin{aligned}
    &\nabla_\lambda\uk=\left(\sum_{i=1}^k{M\left( i,k \right) \odot g_i^a} \right)^T,\\
    &M( i,k) =\sum_{j=i}^k{\frac{\mu _1\left( i,j \right)}{\sqrt{v_j}}-2\text{g}_i\frac{\mu _2\left( i,j \right) m_j}{\sqrt[3]{v_j}}},\\
    &\mu _1\left( i,j \right) =\left( 1-\beta _1 \right) \beta _1^{j-i},\\
    &\mu _2\left( i,j \right) =\left( 1-\beta _2 \right) \beta _{2}^{j-i},
\end{aligned}
\end{equation}
\noindent where $m_i$ and $v_i$ are exponentially averaged first and second order gradient momentum of Adam, and $\beta_1$ and $\beta_2$ are corresponding exponential weights. Theoretical considerations regarding this algorithm and the derivation of Eq.~\eqref{eq:auxgrad2} and Eq.~\eqref{eq:auxgrad3} are given in Appendix A.

We first turn to the optimization problem in \Eqref{object2}. Directly optimizing \Eqref{object2} through gradient-based process needs to save $K$ steps of computational graph and differentiate through the optimization process over $\The$, and it is impractical for \ip. We solve this by computing the explicit form of gradients of \Eqref{object2} and propose the surrogate object function to optimize auxiliary parameters, and the surrogate object function $\tilde{\L}_s(\phi)$ takes the form:
In the remaining part of this paper, we use $\omega$ to denote the function $\omega(\phi)$ to simplify the derivation of our algorithm. 
 the gradient of $\L_s$ respect to $\phi$ takes the form:
\begin{equation}
\label{eq:auxgrad}
\begin{aligned}
\de\phi\is(\The^K(\omega))=&\note{\de\The\L_s(\The^K)}{|1|\times|\theta|} \note{J_{\The^K}(\phi)}{|\The|\times|\phi|} \\
=&\detdels (-\alpha\sum_j (J_a(\The^j))^T J_\omega(\phi))\\
(\textrm{Using } &J_{\The^K}(\phi)=-\alpha\sum_j J_a(\The^j)^T J_\omega(\phi))\\
=& -\alpha \detdels \sum_j J_a(\The^j)^T J_\omega(\phi),\
\end{aligned}
\end{equation}
and the surrogate loss $\tilde{\L_{s}}$ which has the same gradient direction as $\tilde{\L_{s}}$ at point $\omega$ can be defined as:
\begin{equation}
\begin{aligned}
    \tilde{\L_{s}}(\phi)&=\omega\cdot[-\detdels\sum_j J_a(\The^j)^T],\\
    &=-\detdels\sum_j J_a(\The^j)^T\omega.
\end{aligned}
\end{equation}

\ifx\allfiles\undefined
\bibliographystyle{IEEEtran}
\bibliography{reference/reference.bib,reference/diffusion_based.bib,reference/patch_based.bib,reference/diffusion_patch_based.bib,reference/loss_reweighting_alg.bib}